%% file: paper.tex
\documentclass[runningheads]{llncs}
\usepackage{graphicx}
\usepackage{comment}
\usepackage{amsmath,amssymb} 
\usepackage{color}

\input{plots}

\usepackage{hyperref}       
\usepackage{url}            
\usepackage{booktabs}       
\usepackage{amsfonts}       
\usepackage{nicefrac}       
\usepackage{amssymb}
\usepackage{amsmath}
\usepackage{multirow}
\usepackage{enumitem}

\usepackage{algorithm}
\usepackage[noend]{algpseudocode}

\usepackage{tikz}
\usetikzlibrary{shapes.geometric}
\usepackage{rotating}
\usetikzlibrary{positioning}

\usepackage{xcolor}
\usepackage{xspace}

\begin{document}
\pagestyle{headings}
\mainmatter
\def\ECCVSubNumber{1060}  

\input{abbrev}

\title{Graph convolutional networks for learning with few clean and many noisy labels} 

\titlerunning{Graph convolutional networks for learning with few clean and many noisy labels}
%
\author{Ahmet Iscen\inst{1}\and
Giorgos Tolias\inst{2}\and
Yannis Avrithis\inst{3}\and
Ond{\v r}ej Chum\inst{2}\and
Cordelia Schmid\inst{1}
}
\authorrunning{A. Iscen et al.}
%
\institute{Google Research \and
VRG, Faculty of Electrical Engineering, Czech Technical University in Prague \and
Inria, Univ Rennes, CNRS, IRISA
}

\maketitle

\begin{abstract}

In this work we consider the problem of learning a classifier from noisy labels when a few clean labeled examples are given.
The structure of clean and noisy data is modeled by a graph per class and Graph Convolutional Networks (GCN) are used to predict class relevance of noisy examples. For each class, the GCN is treated as a binary classifier, which learns to discriminate clean from noisy examples using a weighted binary cross-entropy loss function. The GCN-inferred ``clean'' probability is then exploited as a relevance measure.
Each noisy example is weighted by its relevance when learning a classifier for the end task.

We evaluate our method on an extended version of a few-shot learning problem, where the few clean examples of novel classes are supplemented with additional noisy data.
Experimental results show that our GCN-based cleaning process significantly improves the classification accuracy over not cleaning the noisy data, as well as standard few-shot classification where only few clean examples are used.

\end{abstract}

\input{intro}
\input{related}
\input{method}
\input{experiments}
\input{conclusions}

\footnotesize{
\head{Acknowledgements.} This work is funded by MSMT LL1901 ERC-CZ grant and OP VVV funded project CZ.02.1.01/0.0/0.0/16\_019/0000765 ``Research Center for Informatics''.
}

\clearpage
{\small
\bibliographystyle{splncs04}
\bibliography{egbib}
}

\input{appendix}

\end{document}

%% file: plots.tex
\usepackage{tikz}
\usetikzlibrary{arrows,shapes,calc,matrix,fit,backgrounds}
\usepackage{pgfplots}
\usepackage{pgfplotstable}
\pgfplotsset{compat=1.9}

\usepackage{xstring}

\usepgfplotslibrary{external}

\newcommand{\extdata}[1]{\input{#1}}
\IfBeginWith*{\jobname}{fig/extern/}{\finalcopy}{}

\newcommand{\leg}[1]{\addlegendentry{#1}}

\tikzset{every mark/.append style={solid}}
\pgfplotsset{
	grid=both, width=\columnwidth, try min ticks=5,
	every axis/.append style={font=\scriptsize},
	every axis plot/.append style={thick,mark=none,mark size=1.2,tension=0.18},
	legend cell align=left, legend style={fill opacity=0.8},
}

\pgfplotsset{
	dash/.style={mark=o,dashed,opacity=0.7},
	dott/.style={mark=o,dotted,opacity=0.7},
}

%% file: abbrev.tex
\newcommand{\nn}[1]{\ensuremath{\text{NN}_{#1}}\xspace}

\newcommand{\citemiss}{\alert{[??]}\xspace}

\newcommand{\supe}[1]{^{\mkern-2mu(#1)}}
\newcommand{\prm}[1]{_{#1}}
\newcommand{\dime}[1]{(#1)}

\def\l1{\ensuremath{\ell_1}\xspace}
\def\l2{\ensuremath{\ell_2}\xspace}

\newcommand*\OK{\ding{51}}

\newenvironment{narrow}[1][1pt]
	{\setlength{\tabcolsep}{#1}}
	{\setlength{\tabcolsep}{6pt}}

\newcommand{\algocom} [1]{{\color{orange} \Comment     #1}} 
\newcommand{\commentout}[1]{}

\newcommand{\alert}[1]{{\color{red}{#1}}}
\newcommand{\head}[1]{{\smallskip\noindent\bf #1}}
\newcommand{\equ}[1]{(\ref{equ:#1})\xspace}

\newcommand{\red}[1]{{\color{red}{#1}}}
\newcommand{\blue}[1]{{\color{blue}{#1}}}
\newcommand{\green}[1]{{\color{green}{#1}}}
\newcommand{\gray}[1]{{\color{gray}{#1}}}


\newcommand{\tran}{^\top}
\newcommand{\mtran}{^{-\top}}
\newcommand{\zcol}{\mathbf{0}}
\newcommand{\zrow}{\zcol\tran}

\newcommand{\ind}{\mathbbm{1}}
\newcommand{\expect}{\mathbb{E}}
\newcommand{\nat}{\mathbb{N}}
\newcommand{\zahl}{\mathbb{Z}}
\newcommand{\real}{\mathbb{R}}
\newcommand{\proj}{\mathbb{P}}
\newcommand{\prob}{\mathbf{Pr}}

\newcommand{\mif}{\textrm{if }}
\newcommand{\other}{\textrm{otherwise}}
\newcommand{\minimize}{\textrm{minimize }}
\newcommand{\maximize}{\textrm{maximize }}
\newcommand{\st}{\textrm{subject to }}

\newcommand{\id}{\operatorname{id}}
\newcommand{\const}{\operatorname{const}}
\newcommand{\sgn}{\operatorname{sgn}}
\newcommand{\var}{\operatorname{Var}}
\newcommand{\mean}{\operatorname{mean}}
\newcommand{\trace}{\operatorname{tr}}
\newcommand{\diag}{\operatorname{diag}}
\newcommand{\vect}{\operatorname{vec}}
\newcommand{\cov}{\operatorname{cov}}

\newcommand{\softmax}{\operatorname{softmax}}
\newcommand{\clip}{\operatorname{clip}}

\newcommand{\defn}{\mathrel{:=}}
\newcommand{\peq}{\mathrel{+\!=}}
\newcommand{\meq}{\mathrel{-\!=}}

\newcommand{\floor}[1]{\left\lfloor{#1}\right\rfloor}
\newcommand{\ceil}[1]{\left\lceil{#1}\right\rceil}
\newcommand{\inner}[1]{\left\langle{#1}\right\rangle}
\newcommand{\norm}[1]{\left\|{#1}\right\|}
\newcommand{\frob}[1]{\norm{#1}_F}
\newcommand{\card}[1]{\left|{#1}\right|\xspace}
\newcommand{\diff}{\mathrm{d}}
\newcommand{\der}[3][]{\frac{d^{#1}#2}{d#3^{#1}}}
\newcommand{\pder}[3][]{\frac{\partial^{#1}{#2}}{\partial{#3^{#1}}}}
\newcommand{\ipder}[3][]{\partial^{#1}{#2}/\partial{#3^{#1}}}
\newcommand{\dder}[3]{\frac{\partial^2{#1}}{\partial{#2}\partial{#3}}}

\newcommand{\wb}[1]{\overline{#1}}
\newcommand{\wt}[1]{\widetilde{#1}}

\def\xssp{\hspace{1pt}}
\def\ssp{\hspace{3pt}}
\def\msp{\hspace{5pt}}
\def\lsp{\hspace{12pt}}

\newcommand{\cA}{\mathcal{A}}
\newcommand{\cB}{\mathcal{B}}
\newcommand{\cC}{\mathcal{C}}
\newcommand{\cD}{\mathcal{D}}
\newcommand{\cE}{\mathcal{E}}
\newcommand{\cF}{\mathcal{F}}
\newcommand{\cG}{\mathcal{G}}
\newcommand{\cH}{\mathcal{H}}
\newcommand{\cI}{\mathcal{I}}
\newcommand{\cJ}{\mathcal{J}}
\newcommand{\cK}{\mathcal{K}}
\newcommand{\cL}{\mathcal{L}}
\newcommand{\cM}{\mathcal{M}}
\newcommand{\cN}{\mathcal{N}}
\newcommand{\cO}{\mathcal{O}}
\newcommand{\cP}{\mathcal{P}}
\newcommand{\cQ}{\mathcal{Q}}
\newcommand{\cR}{\mathcal{R}}
\newcommand{\cS}{\mathcal{S}}
\newcommand{\cT}{\mathcal{T}}
\newcommand{\cU}{\mathcal{U}}
\newcommand{\cV}{\mathcal{V}}
\newcommand{\cW}{\mathcal{W}}
\newcommand{\cX}{\mathcal{X}}
\newcommand{\cY}{\mathcal{Y}}
\newcommand{\cZ}{\mathcal{Z}}

\newcommand{\vA}{\mathbf{A}}
\newcommand{\vB}{\mathbf{B}}
\newcommand{\vC}{\mathbf{C}}
\newcommand{\vD}{\mathbf{D}}
\newcommand{\vE}{\mathbf{E}}
\newcommand{\vF}{\mathbf{F}}
\newcommand{\vG}{\mathbf{G}}
\newcommand{\vH}{\mathbf{H}}
\newcommand{\vI}{\mathbf{I}}
\newcommand{\vJ}{\mathbf{J}}
\newcommand{\vK}{\mathbf{K}}
\newcommand{\vL}{\mathbf{L}}
\newcommand{\vM}{\mathbf{M}}
\newcommand{\vN}{\mathbf{N}}
\newcommand{\vO}{\mathbf{O}}
\newcommand{\vP}{\mathbf{P}}
\newcommand{\vQ}{\mathbf{Q}}
\newcommand{\vR}{\mathbf{R}}
\newcommand{\vS}{\mathbf{S}}
\newcommand{\vT}{\mathbf{T}}
\newcommand{\vU}{\mathbf{U}}
\newcommand{\vV}{\mathbf{V}}
\newcommand{\vW}{\mathbf{W}}
\newcommand{\vX}{\mathbf{X}}
\newcommand{\vY}{\mathbf{Y}}
\newcommand{\vZ}{\mathbf{Z}}

\newcommand{\va}{\mathbf{a}}
\newcommand{\vb}{\mathbf{b}}
\newcommand{\vc}{\mathbf{c}}
\newcommand{\vd}{\mathbf{d}}
\newcommand{\ve}{\mathbf{e}}
\newcommand{\vf}{\mathbf{f}}
\newcommand{\vg}{\mathbf{g}}
\newcommand{\vh}{\mathbf{h}}
\newcommand{\vi}{\mathbf{i}}
\newcommand{\vj}{\mathbf{j}}
\newcommand{\vk}{\mathbf{k}}
\newcommand{\vl}{\mathbf{l}}
\newcommand{\vm}{\mathbf{m}}
\newcommand{\vn}{\mathbf{n}}
\newcommand{\vo}{\mathbf{o}}
\newcommand{\vp}{\mathbf{p}}
\newcommand{\vq}{\mathbf{q}}
\newcommand{\vr}{\mathbf{r}}
\newcommand{\vs}{\mathbf{s}}
\newcommand{\vt}{\mathbf{t}}
\newcommand{\vu}{\mathbf{u}}
\newcommand{\vv}{\mathbf{v}}
\newcommand{\vw}{\mathbf{w}}
\newcommand{\vx}{\mathbf{x}}
\newcommand{\vy}{\mathbf{y}}
\newcommand{\vz}{\mathbf{z}}

\newcommand{\vone}{\mathbf{1}}
\newcommand{\vzero}{\mathbf{0}}

\newcommand{\valpha}{{\boldsymbol{\alpha}}}
\newcommand{\vbeta}{{\boldsymbol{\beta}}}
\newcommand{\vgamma}{{\boldsymbol{\gamma}}}
\newcommand{\vdelta}{{\boldsymbol{\delta}}}
\newcommand{\vepsilon}{{\boldsymbol{\epsilon}}}
\newcommand{\vzeta}{{\boldsymbol{\zeta}}}
\newcommand{\veta}{{\boldsymbol{\eta}}}
\newcommand{\vtheta}{{\boldsymbol{\theta}}}
\newcommand{\viota}{{\boldsymbol{\iota}}}
\newcommand{\vkappa}{{\boldsymbol{\kappa}}}
\newcommand{\vlambda}{{\boldsymbol{\lambda}}}
\newcommand{\vmu}{{\boldsymbol{\mu}}}
\newcommand{\vnu}{{\boldsymbol{\nu}}}
\newcommand{\vxi}{{\boldsymbol{\xi}}}
\newcommand{\vomikron}{{\boldsymbol{\omikron}}}
\newcommand{\vpi}{{\boldsymbol{\pi}}}
\newcommand{\vrho}{{\boldsymbol{\rho}}}
\newcommand{\vsigma}{{\boldsymbol{\sigma}}}
\newcommand{\vtau}{{\boldsymbol{\tau}}}
\newcommand{\vupsilon}{{\boldsymbol{\upsilon}}}
\newcommand{\vphi}{{\boldsymbol{\phi}}}
\newcommand{\vchi}{{\boldsymbol{\chi}}}
\newcommand{\vpsi}{{\boldsymbol{\psi}}}
\newcommand{\vomega}{{\boldsymbol{\omega}}}

\newcommand{\rLambda}{\mathrm{\Lambda}}
\newcommand{\rSigma}{\mathrm{\Sigma}}

\def\onedot{.\xspace}
\def\eg{\emph{e.g}\onedot} \def\Eg{\emph{E.g}\onedot}
\def\ie{\emph{i.e}\onedot} \def\Ie{\emph{I.e}\onedot}
\def\cf{\emph{cf}\onedot} \def\Cf{\emph{C.f}\onedot}
\def\etc{\emph{etc}\onedot}
\def\vs{\emph{vs}\onedot}
\def\wrt{w.r.t\onedot} \def\dof{d.o.f\onedot}
\def\etal{\emph{et al}.}

\newcommand{\std}[1]{\tiny{$\pm$#1}}

\makeatother

%% file: intro.tex
\section{Introduction}
\label{sec:intro}

State-of-the-art deep learning methods require a large amount of manually labeled data. The need for supervision may be reduced by decoupling representation learning from the end task and/or using additional training data that is unlabeled, weakly labeled (with noisy labels), or belong to different domains or classes. Example approaches are \emph{transfer learning}~\cite{WG15}, \emph{unsupervised representation learning}~\cite{WG15}, \emph{semi-supervised learning}~\cite{WeRC08}, \emph{learning from noisy labels}~\cite{joulin2016learning} and \emph{few-shot learning}~\cite{SSZ17}.

However, for 
several classes,
only very few or even no clean labeled examples might be available at the representation learning stage.
\emph{Few-shot learning} 
severely limits the number of labeled samples on the end task,
while the representation is learned on a large training set of different classes~\cite{HG17,SSZ17,VBL+16}. \commentout{Nevertheless, what if more data with noisy labels is actually available on the end task?}
Nevertheless, in many situations, more data with noisy labels 
can be acquired or is readily available for the end task.

One interesting mix of few-shot learning with additional large-scale data is the work of Douze~\etal~\cite{DSH+18}, where labels are propagated from few clean labeled examples to a large-scale collection.
This collection is unlabeled and actually contains data for many more classes than the end task.
Their method overall improves the classification accuracy, but at an additional computational cost.
It is a \emph{transductive} method, 
\ie, instead of learning a parametric classifier, the large-scale collection is still necessary at inference.

\begin{figure}[t]
\centering
\includegraphics[width=0.99\linewidth]{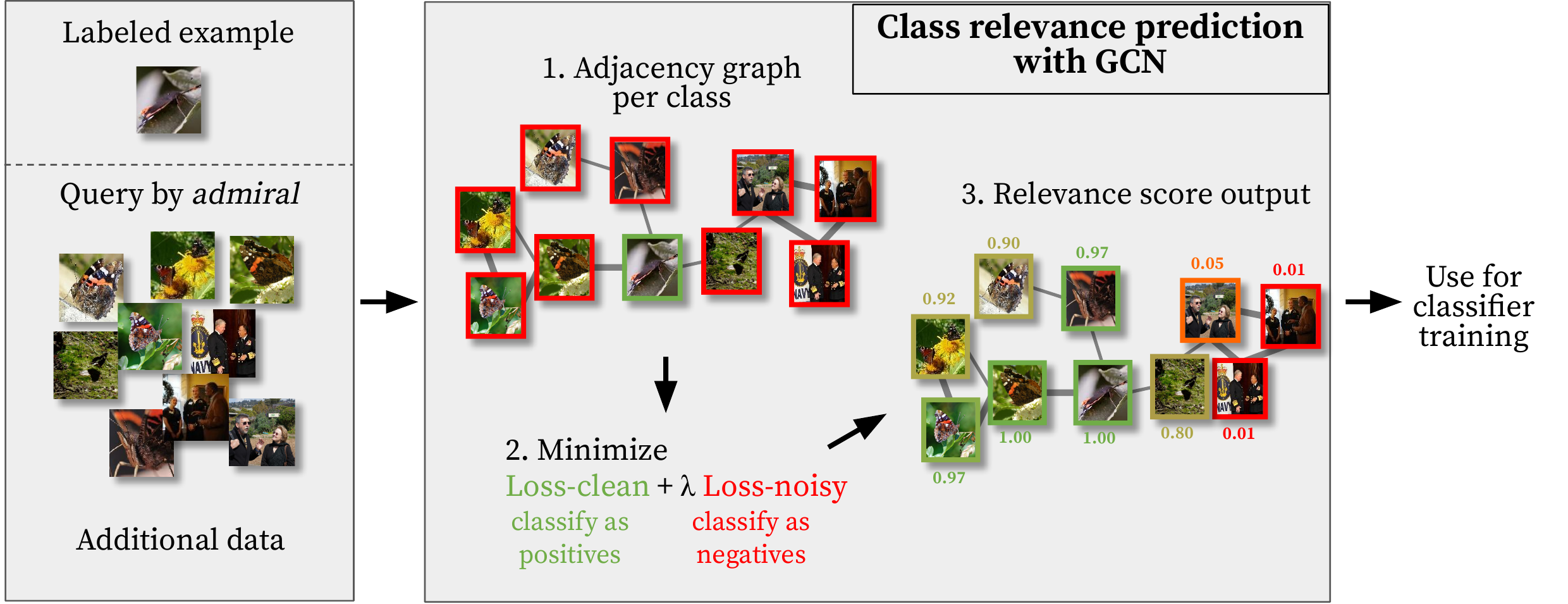}
\caption{Overview of our cleaning approach for 1-shot learning  with noisy examples. We use the class name \emph{admiral} to crawl noisy images from the web and create an adjacency graph based on visual similarity. We then assign a relevance score to each noisy example with a graph convolutional network (GCN).  Relevance scores are displayed next to the images.\label{fig:teaser}
}
\end{figure}

In this work, we learn a classifier from a few clean labeled examples and additional weakly labeled data, while the representation is learned on different classes, 
similarly to
few-shot learning. We assume that the class names are known, and we use them to search an existing large collection of images with textual description. The result is a
set of images with novel class labels, but potentially incorrect (noisy).
As shown in~ \autoref{fig:teaser}, we clean this data using a \emph{graph convolutional network} (GCN)~\cite{KW17}, which learns to predict a class relevance score per image based on connections to clean images in the graph.
Both the clean and the noisy images are then used to learn a classifier, 
where the noisy examples weighted by relevance.
Unlike most existing work, our method operates independently per class and applies when clean labeled examples are few or even only one per class.

We make the following contributions:
\begin{enumerate}[noitemsep,topsep=0pt]
	\item We learn a classifier on a \emph{large-scale weakly-labeled} collection jointly with only a \emph{few clean labeled} examples.
	\item To our knowledge, we are the first to use a GCN to clean noisy data: we cast a GCN as a \emph{binary classifier} which learns to discriminate clean from noisy data, and we use its inferred ``clean'' probabilities as a relevance score per example.
	\item We apply our method to two few-shot learning benchmarks and show significant improvement in accuracy, outperforming the method by Douze~\etal~\cite{DSH+18} using the same large-scale collection of data without labels.
\end{enumerate}

%% file: related.tex
\section{Related work}
\label{sec:related}

\head{Learning with noisy labels} is often concerned with estimating or learning a \emph{transition matrix}~\cite{NDRT13,PRK+17,sukhbaatar2014training} or \emph{knowledge graph}~\cite{li2017learning} between labels and correcting the loss function, which does not apply in our case since the true classes in the noisy data is unknown. Most recent work on learning from large-scale weakly-labeled data focuses on learning the \emph{representation} 
through
\emph{metric learning}~\cite{lee2018cleannet,wang2018iterative},
\emph{bootstrapping}~\cite{reed2014training}, or \emph{distillation}~\cite{li2017learning}.
In our case however, since the clean labeled examples are 
scarce,
we need to keep the representation mostly fixed.

\emph{Dealing with the noise} by thresholding~\cite{lee2018cleannet}, outlier detection~\cite{wang2018iterative} or \emph{reweighting}~\cite{liu2015classification}, is applicable while the representation is learned, based on the gradient of the loss~\cite{ren2018learning}. In contrast,
the relatively-shallow GCN that we propose effectively decouples reweighting from
both representation learning and classifier learning.
\emph{Learning to clean} the noisy labels~\cite{Veit_2017_CVPR} typically assumes adequate human verified labels for training, which again is
not the case in this work.

\head{Few-shot learning.}
\emph{Meta-learning}~\cite{vilalta2002perspective} refers to learning at two levels, where generic knowledge is acquired before adapting to more specific tasks. In few-shot learning, this translates to learning on a set of \emph{base classes} how to learn from few examples on a distinct set of \emph{novel classes} without overfitting. For instance, \emph{optimization meta-learning}~\cite{finn2017model,FXL18,RL16} amounts to learning a model that is easy to fine-tune in few steps. In our work, we study an extension of few-shot learning where more 
novel class instances are available, reducing the risk of overfitting when fine-tuning the model. \emph{Metric learning} approaches learn how to compare queries for instance to few examples~\cite{VBL+16} or to the corresponding \emph{class prototypes}~\cite{SSZ17}. Hariharan and Girshick~\cite{HG17} and Wang~\etal~\cite{WGH+18} learn how to \emph{generate} novel-class examples,
which is not needed when more data is actually available.

Gidaris and Komodakis~\cite{GK18} learn on base classes a simpler cosine similarity-based parametric classifier, or simply \emph{cosine classifier}, without meta-learning.
The same classifier has been introduced independently by Qi~\etal~\cite{Qi_2018_CVPR}, who further fine-tune the network, assuming access to the base class training set.
A recent survey~\cite{CLK+19} confirms the superiority of the cosine classifier to previous work including meta-learning~\cite{finn2017model}. We use the cosine classifier in this work, both for base and novel classes.

Making use of \emph{unlabeled data} has been little explored in few-shot learning until recently.
Ren~\etal~\cite{RRTSST18} introduce a semi-supervised few-shot classification task, where some labels are unknown.
Liu~\etal~\cite{liu2018learning} follow the same semi-supervised setup, but use graph-based \emph{label propagation} (LP)~\cite{ZBL+03} for classification and consider 
all test images jointly.
These methods assume a meta-learning scenario,
where only small-scale data is available at each
training episode; arguably, such a small amount of data limits the representation adaptation and generalization to unseen data. Similarly, Rohrbach~\etal~\cite{RES13} use label propagation in a \emph{transductive} setting, but at a larger scale assuming that all examples come from a set of known classes.
Douze~\etal~\cite{DSH+18} extend to an even larger scale, leveraging 100M unlabeled images in a graph without using additional text information. We focus on the latter large-scale scenario using the same 100M dataset. However, we filter by text to obtain \emph{noisy data} and follow an \emph{inductive} approach by training a classifier for novel classes, such that the 100M collection is not needed at inference.

\head{Graph neural networks} are generalizations of convolutional networks to non-Euclidean spaces~\cite{BBL+16}. Early \emph{spectral methods}~\cite{BZSL13,HeBL15} have been succeeded by \emph{Chebyshev polynomial} approximations~\cite{DeBV16}, which avoid the high computational cost of computing eigenvectors. \emph{Graph convolutional networks} (GCN)~\cite{KW17} provide a further simplification by a \emph{first-order} approximation of graph filtering and are applied to \emph{semi-supervised}~\cite{KW17} and subsequently \emph{few-shot learning}~\cite{GaBr18}.
Kipf and Welling~\cite{KW17} apply the loss function to labeled examples to make predictions on unlabeled ones. Similarly, Garcia and Bruna~\cite{GaBr18} use GCNs to make predictions on novel class examples.
Gidaris and Komodakis~\cite{GK19} use Graph Neural Networks as denoising autoencoders to generate class weights for novel classes.
By contrast, we cast GCNs as \emph{binary classifiers} discriminating clean from noisy examples: we apply a loss function to all examples, and then use the inferred probabilities as a class relevance measure, effectively cleaning the data.

Our counter-intuitive objective of treating all noisy examples as negative
can be compared to treating each example as a different class in \emph{instance-level discrimination}~\cite{wu2018unsupervised}.
In fact, our loss function is similar to \emph{noise-contrastive estimation} (NCE)~\cite{GuHy10}.
Our experiments show that our GCN-based classifier
outperforms classical LP~\cite{ZBL+03} used for a similar purpose by~\cite{RES13}.

%% file: method.tex
\section{Problem formulation}
\label{sec:problem}

We consider a space $\cX$ of examples.
We are given a set $X_\cC \subset \cX$ of examples, each having a \emph{clean} (manually verified) label in a set $C$ of classes with $\card{C}=K$.
We assume that the number $\card{X_\cC^c}$ of examples\footnote{For any set $X \subset \cX$, we denote by $X^c$ its subset of examples labeled in class $c \in C$.} labeled in each class $c \in C$ is only $k$, typically in $\{1,2,5,10,20\}$.
We are also given an additional set $X_\cN$ of examples, each with a set of \emph{noisy} labels in $C$.
The \emph{extended} set of examples for class $c$ is now $X_\cE^c = X_\cC^c \cup X_\cN^c$.
Examples or sets of examples having clean (noisy) labels are referred to as clean (noisy) as well.
The goal is to train a $K$-way classifier, using the additional noisy set in order to improve the accuracy compared to only using the small clean set.

We assume that we are given a feature extractor $g_\theta: \cX \to \real^d$, mapping an example to a $d$-dimensional vector. For instance, when examples are images, the feature extractor is typically a \emph{convolutional neural network} (CNN) and $\theta$ are the parameters of all layers.

In this work, we assume that the noisy set $X_\cN$ is collected via web crawling. Examples are images accompanied by free-form text descriptions and/or user tags originating from community photo collections. To make use of text data, we assume that the names of the classes in $C$ are given.
An example in $X_\cN$ is given a label in class $c \in C$ if its textual information contains the name of class $c$; it may then have none, one or more labels.
In this way, we automatically infer labels for $X_\cN$ without human effort, which are however \emph{noisy}.

\section{Cleaning with graph convolutional networks}
\label{sec:gcn}

We perform cleaning by predicting a \emph{class relevance} measure for each noisy example in $X_\cN^c$, independently 
for each 
class $c \in C$. To simplify the notation, we drop superscript $c$ where possible in this subsection and we denote $X_\cE^c$ by $\{x_1, \dots, \allowbreak x_k, x_{k+1}, \dots, x_{N}\}$, where $X_\cC^c = \{x_1, \dots, x_k\}$ and $X_\cN^c = \{x_{k+1}, \dots, x_{N}\}$. The features of these examples are similarly represented by matrix $V = [\vv_1, \dots,\allowbreak \vv_k, \vv_{k+1}, \dots, \vv_N] \in \real^{d \times N}$, where $\vv_i = g_\theta(x_i)$ for $i=1,\dots,N$.

We construct an affinity matrix $A \in \real^{N \times N}$ with elements $a_{ij} = [\vv_i^\top \vv_j]_+$ if examples $\vv_i$ and $\vv_j$ are reciprocal nearest neighbors in $X_\cE^c$ and $0$ otherwise. Matrix $A$ has zero diagonal, but self-connections are added 
before
$A$ is normalized as $\tilde{A} = D^{-1}(A + I)$ with $D = \diag((A+I) \vone)$ being the degree matrix of $A+I$ and $\vone$ the all-ones vector.

\emph{Graph convolutional networks} (GCNs)~\cite{KW17} are formed by a sequence of layers. Each layer is a function $f_{\Theta}: \real^ {N \times N} \times \real^ {l \times N} \to \real^ {n \times N}$ of the form
\begin{equation}
f_{\Theta}(\tilde{A},Z) = h(\Theta\tran Z \tilde{A}),
\label{equ:gcnlayer}
\end{equation}
where $Z \in \real^{l \times N}$ represents the input features, $\Theta \in \real^{l \times n}$ holds the parameters of the layer to be learned, and $h$ is a nonlinear activation function. Function $f_{\Theta}$ maps $l$-dimensional input features to $n$-dimensional output features.

In this work we consider a two-layer GCN with a scalar output per example. This network is a function $F_{\Theta}: \real^{N \times N} \times \real^{d \times N} \to \real^N$ given by
\begin{align}
F_{\Theta}(\tilde{A},V) &= \sigma(\Theta_2\tran [\Theta_1\tran V \tilde{A}]_{+} \tilde{A}),
\label{equ:gcn}
\end{align}
where $\Theta = \{\Theta_1, \Theta_2\}$, $\Theta_1 \in \real^{d \times m}$, $\Theta_2 \in \real^{m \times 1}$, $[\cdot]_+$ is the positive part or ReLU function~\cite{NaHi10} and $\sigma(a) = (1+e^{-a})^{-1}$ for $a \in \real$ is the sigmoid function. Function $F_{\Theta}$ performs feature propagation through the affinity matrix in an analogy to classical graph-based propagation methods for classification~\cite{ZBL+03} or search~\cite{ZWG+03}.

The output $F_{\Theta}(\tilde{A},V)$ is a vector of length $N$, with element $F_{\Theta}(\tilde{A},V)_i$ in $[0,1]$ representing a relevance value of example $x_i$ for class $c$. To learn the parameters $\Theta$, we treat the GCN as a \emph{binary classifier} where target output $1$ corresponds to clean examples and $0$ to noisy. In particular, we minimize the loss function
\vspace{-5pt}
\begin{align}
\hspace{-8pt}
L_\cG(V, \tilde{A}; \Theta) =
	-\frac{1}{k} \sum_{i=1}^k \log \left( F_{\Theta}(\tilde{A},V)_i \right)
	- \frac{\lambda}{N-k} \sum_{i=k+1}^N \log \left( 1-F_{\Theta}(\tilde{A},V)_i \right).
\label{equ:gcn-loss}
\end{align}
This is a binary cross-entropy loss function where noisy examples are given an importance weight $\lambda$. Given the propagation on the nearest neighbor graph, and depending on the relative importance $\lambda$ of the second term, noisy examples that are strongly connected to clean ones are still expected to receive high class relevance, while noisy examples that are not relevant to the current class are expected to get a class relevance near zero.

The impact of parameter $\lambda$ is validated in Section~\ref{sec:exp}, where we show that the fewer the available clean images are (smaller $k$) the smaller the importance weight should be.
As is standard practice for GCNs in classification~\cite{KW17}, training is performed in batches of size $N$, that is the entire set of features.

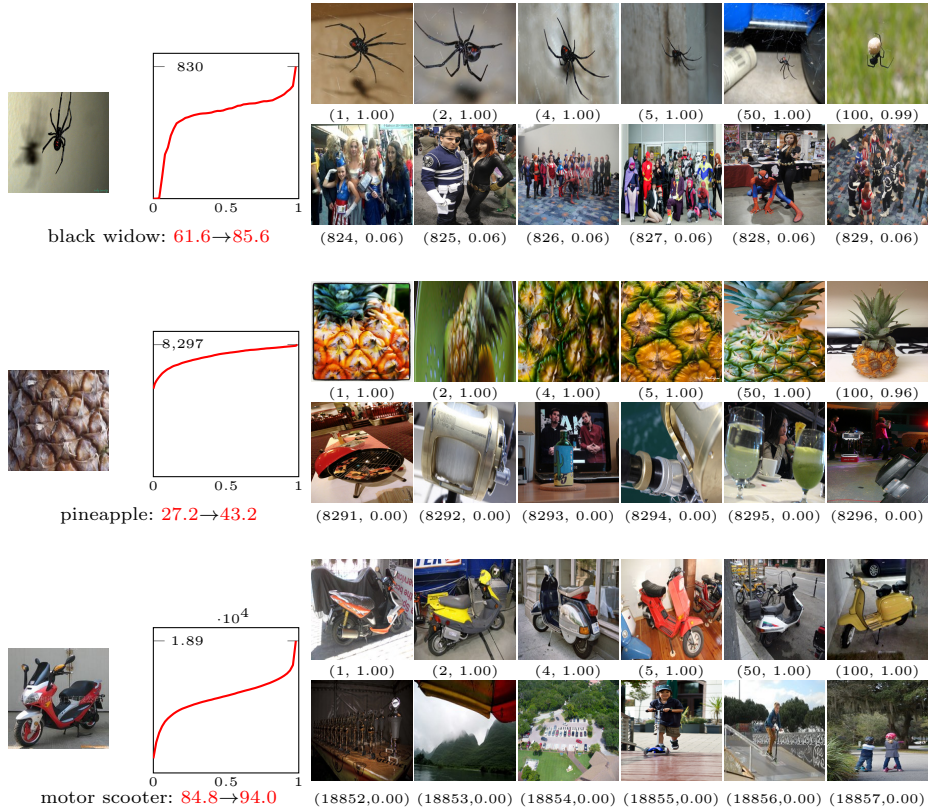
\begin{figure}[t]
\input{fig_example2}

\caption{Examples of clean images from the Low-Shot ImageNet Benchmark (left) for 1-shot classification, cumulative histogram of the predicted relevance for noisy images (middle) and representative noisy images (right)
by descending order of
relevance, with relevance value reported below. Test accuracy without and with additional data using class prototypes~\equ{proto} is shown next to class names.
\label{fig:example2}
}
\end{figure}

Figure~\ref{fig:example2} shows examples of clean images, corresponding noisy ones and the predicted relevance.
Using the visual similarity to the clean image, we can use relevance to resolve cases of polysemy, \eg \emph{black widow (spider)} \vs \emph{black widow (superhero)}, or cases like \emph{pineapple} \vs \emph{pineapple juice}.

\head{Discussion.} Loss function~\equ{gcn-loss} is similar to \emph{noise-contrastive estimation} (NCE) \cite{GuHy10} as used by We \etal~\cite{wu2018unsupervised} for \emph{instance-level discrimination}, whereas we discriminate clean from noisy examples.
The semi-supervised learning setup of GCNs~\cite{KW17} uses a loss function that applies only to the labeled examples, and makes discrete predictions on unlabeled examples. In our case, all examples contribute to the loss but with different importance, 
as
we infer real-valued class relevance for the noisy examples, to be used for subsequent learning.

Function $F_{\Theta}$ in~\equ{gcn} reduces to a Multi-Layer Perceptron (MLP) when the affinity matrix $A$ is zero, in which case all examples are disconnected. Using an MLP to perform cleaning would take each example into account independently of the others, while the GCN considers the collection of examples as a whole.
MLP training is performed identically to GCN by minimizing~\equ{gcn-loss}.
We compare the two alternatives in our experiments.

\section{Learning a classifier with few clean and many noisy examples}
\label{sec:evalproto}
Our cleaning process applies when the clean labeled examples are few, but assumes a feature extractor\footnote{For instance, after training on a different task or a set of classes other than $C$. Learning of the feature extractor used in our experiments is described in Appendix.} $g_\theta$. That is, representation learning, label cleaning and classifier learning are decoupled.
We perform GCN-based cleaning as described in Section~\ref{sec:gcn}, and learn a classifier by weighting examples according to class relevance.
The process of training the classifier
is described below.

\subsection{Cosine-similarity based classifier}
\label{sec:cosclassifier}
We use a \emph{cosine-similarity based classifier}~\cite{GK18,Qi_2018_CVPR}, or \emph{cosine classifier} for short.
Each class $c \in C$ is represented by a learnable parameter $\vw_c \in \real^d$. The \emph{prediction} of example $x \in \cX$ is the class $c$ of maximum cosine similarity $\hat{\vw}_c\tran \hat{g}_\theta(x)$\footnote{We denote the $\ell_2$-normalized counterpart of vector $\vx$ by $\hat{\vx}$. Similarly, if $\vy = f(x)$, we denote $\hat{\vy}$ by $\hat{f}(x)$.}
\begin{equation}
\pi_{\theta,W}(x) = \arg\max_c \hat{\vw}_c\tran \hat{g}_\theta(x),
\label{equ:classify}
\end{equation}
where $W = [\vw_1, \dots, \vw_K] \in \real^{d\times K}$.

\subsection{Classifier learning}
\label{sec:novelclass}
The goal is to learn a $K$-way classifier for unseen data in $\cX$.
Unlike the typical few-shot learning task, each class contains a few clean and many noisy examples.

Prior to learning classifiers, training examples $x_i \in X_\cE^c$ are weighted by their relevance $r(x_i)$ to class $c$. For a noisy example $x_i \in X_\cN^c$, we define $r(x_i) = F_{\Theta}(\tilde{A},V)_i$ where $F_{\Theta}(\tilde{A},V)$ is the output vector of the GCN, while for a clean example $x_i \in X_\cC^c$ we fix $r(x_i) = 1$.
Note that optimizing~\equ{gcn-loss} does not guarantee $F_{\Theta}(\tilde{A},V)_i = 1$ for clean examples $x_i \in X_\cC^c$.
We define $r(X) = \sum_{x \in X} r(x)$ for any set $X \subset \cX$.

We first assume that the given feature extractor is fixed and consider two different classifiers, namely class prototypes and cosine-similarity based classifier. Then, this assumption is dropped and the classifier and feature representation are learned jointly by fine-tuning the entire network.

\head{Class prototypes.}
For each class $c \in C$, we define \emph{prototype} $\vw_c$ by
\begin{equation}
	\vw_c = \frac{1}{r(X_\cE^c)} \sum_{x \in X_\cE^c} r(x) g_\theta(x).
\label{equ:proto}
\end{equation}
Prototypes are fixed vectors, not learnable parameters. Collecting them into matrix $W = [\vw_1, \dots, \vw_{K}] \in \real^{d\times K}$, $K$-way prediction is made by classifier $\pi_{\theta,W}$~\equ{classify}.

\head{Cosine classifier learning.}
Given examples $X_\cE$, we learn a parametric cosine classifier with parameters $W = [\vw_1, \dots, \vw_{K}] \in \real^{d\times K}$ by minimizing the weighted cross entropy loss $L(C, X_\cE, \theta; W)$ over $W$, given by
\begin{equation}
L(C, X_\cE, \theta; W) = - \sum_{c \in C} \frac{1}{r(X_\cE^c)} \sum_{x \in X_\cE^c}
	r(x) \log(\vsigma(s \hat{W}\tran \hat{g}_\theta(x))_c),
\label{equ:loss}
\end{equation}
where $\vsigma: \real^K \to \real^K$ is the softmax function with $\vsigma(\va)_c = e^{a_c} / \sum_{j \in C} e^{a_j}$ for $\va \in \real^K$, $s$ is a \emph{scale parameter} and $\hat{W} = [\hat{\vw}_1, \dots, \hat{\vw}_K]$.
The parameters $\theta$ of the feature extractor are fixed. The scale parameter $s$ is also fixed according to the training of the feature extractor.
Prediction is made as in the previous case.

\head{Deep network fine-tuning.} An alternative is to drop the assumption that the feature extractor is fixed. In this case, we jointly learn the parameters $\theta$ of the feature extractor and $W$ of the $K$-way cosine classifier by minimizing
the right-hand side of \equ{loss}.
This requires access to examples $X_\cE$, while for the previous two classifiers, access to features $g_\theta(x)$ is enough.
Note that, due to over-fitting on the few available examples, such learning is typically avoided in a few-shot learning setup.

%% file: fig_example2.tex
\tiny
\begin{tabular}{@{\xssp}c@{\hspace{-50pt}}c@{\xssp}c@{\xssp}c@{\xssp}c@{\xssp}c@{\xssp}c@{\xssp}c@{\xssp}c@{\xssp}}
\multirow{2}{*}{\includegraphics[width=38pt,height=38pt]{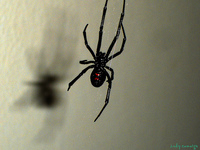}}&
\raisebox{15pt}{
\multirow{2}{*}{\input{fig/data/sample.tex}
\hspace{-15pt}
\begin{tikzpicture}
\begin{axis}[%
    width=100pt,
    height=100pt,
    xmin = 0, xmax = 1,
    xtick style={draw=none},
    ymin = 0,
    grid=none,
    ytick={830},
    xtick = {0,0.5,1},    
    xticklabel style = {yshift=0.3ex,font=\tiny},
    yticklabel style = {xshift=10ex,font=\tiny},
    ylabel style = {yshift=-1.9ex,font=\tiny},
]
\addplot[color=red] table[x=x,y=z] \histclasstwofive;
\end{axis}
\end{tikzpicture}}}&
\includegraphics[width=38pt,height=38pt]{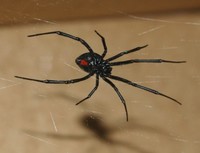}&
\includegraphics[width=38pt,height=38pt]{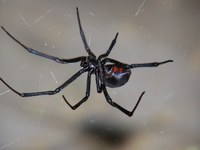}&
\includegraphics[width=38pt,height=38pt]{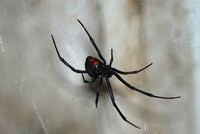}&
\includegraphics[width=38pt,height=38pt]{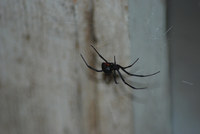}&
\includegraphics[width=38pt,height=38pt]{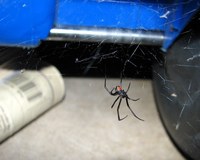}&
\includegraphics[width=38pt,height=38pt]{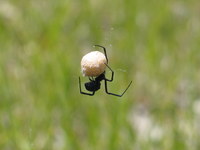}&\\[0pt]
&&(1, 1.00)&(2, 1.00)&(4, 1.00)&(5, 1.00)&(50, 1.00)&(100, 0.99)\\ & &
\includegraphics[width=38pt,height=38pt]{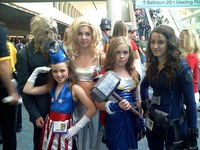}&
\includegraphics[width=38pt,height=38pt]{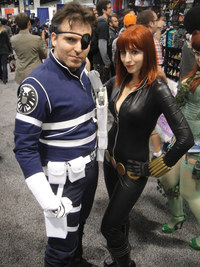}&
\includegraphics[width=38pt,height=38pt]{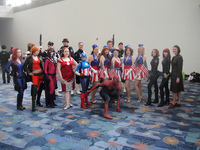}&
\includegraphics[width=38pt,height=38pt]{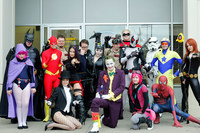}&
\includegraphics[width=38pt,height=38pt]{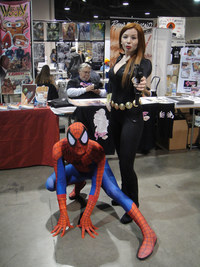}&
\includegraphics[width=38pt,height=38pt]{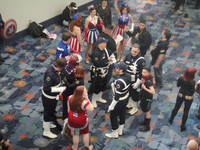}&\\[0pt]
\multicolumn{2}{c}{\scriptsize{black widow: \alert{61.6}$\rightarrow$\alert{85.6}}}&(824, 0.06)&(825, 0.06)&(826, 0.06)&(827, 0.06)&(828, 0.06)&(829, 0.06)\\[13pt]
%
%
\multirow{2}{*}{\includegraphics[width=38pt,height=38pt]{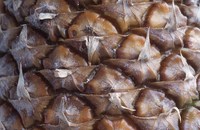}}&
\raisebox{15pt}{
\multirow{2}{*}{\input{fig/data/sample.tex}
\hspace{-15pt}
\begin{tikzpicture}
\begin{axis}[%
    width=100pt,
    height=100pt,
    xmin = 0, xmax = 1,
    xtick style={draw=none},
    ymin = 0,
    grid=none,
    ytick={8297},
    xtick = {0,0.5,1},    
    xticklabel style = {yshift=0.3ex,font=\tiny},
    yticklabel style = {xshift=10ex,font=\tiny},
    ylabel style = {yshift=-1.9ex,font=\tiny},
]
\addplot[color=red] table[x=x,y=z] \histclasstwoninefive;
\end{axis}
\end{tikzpicture}}}&
\includegraphics[width=38pt,height=38pt]{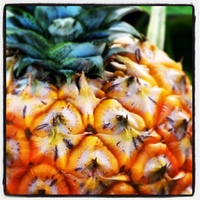}&
\includegraphics[width=38pt,height=38pt]{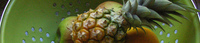}&
\includegraphics[width=38pt,height=38pt]{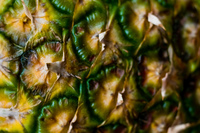}&
\includegraphics[width=38pt,height=38pt]{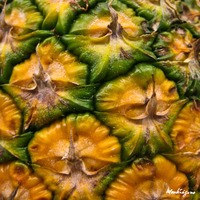}&
\includegraphics[width=38pt,height=38pt]{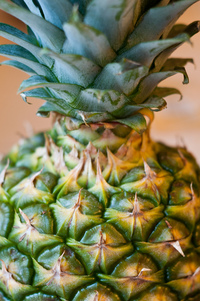}&
\includegraphics[width=38pt,height=38pt]{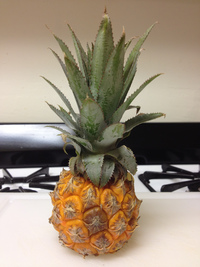}&\\[0pt]
&&(1, 1.00)&(2, 1.00)&(4, 1.00)&(5, 1.00)&(50, 1.00)&(100, 0.96)\\ & &
\includegraphics[width=38pt,height=38pt]{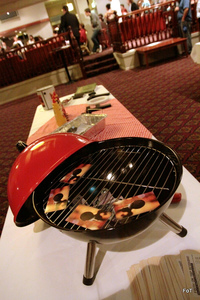}&
\includegraphics[width=38pt,height=38pt]{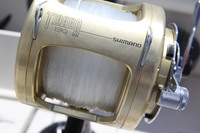}&
\includegraphics[width=38pt,height=38pt]{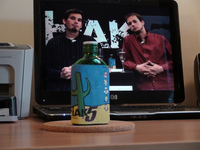}&
\includegraphics[width=38pt,height=38pt]{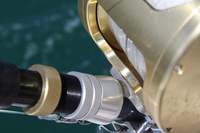}&
\includegraphics[width=38pt,height=38pt]{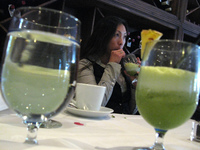}&
\includegraphics[width=38pt,height=38pt]{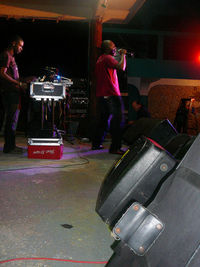}&\\[0pt]
\multicolumn{2}{c}{\scriptsize{pineapple: \alert{27.2}$\rightarrow$\alert{43.2} }}&(8291, 0.00)&(8292, 0.00)&(8293, 0.00)&(8294, 0.00)&(8295, 0.00)&(8296, 0.00)\\[13pt]
%
%
\multirow{2}{*}{\includegraphics[width=38pt,height=38pt]{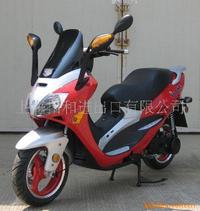}}&
\raisebox{15pt}{
\multirow{2}{*}{\input{fig/data/sample.tex}
\hspace{-15pt}
\begin{tikzpicture}
\begin{axis}[%
    width=100pt,
    height=100pt,
    xmin = 0, xmax = 1,
    xtick style={draw=none},
    ymin = 0,
    grid=none,
    ytick={18858},
    xtick = {0,0.5,1},    
    xticklabel style = {yshift=0.3ex,font=\tiny},
    yticklabel style = {xshift=10ex,font=\tiny},
    ylabel style = {yshift=-1.9ex,font=\tiny},
]
\addplot[color=red] table[x=x,y=z] \histclasstwoonesix;
\end{axis}
\end{tikzpicture}}}&
\includegraphics[width=38pt,height=38pt]{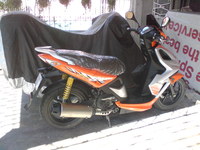}&
\includegraphics[width=38pt,height=38pt]{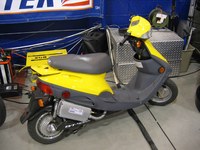}&
\includegraphics[width=38pt,height=38pt]{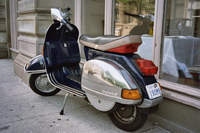}&
\includegraphics[width=38pt,height=38pt]{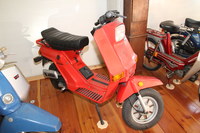}&
\includegraphics[width=38pt,height=38pt]{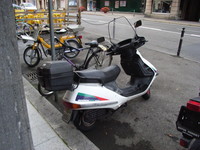}&
\includegraphics[width=38pt,height=38pt]{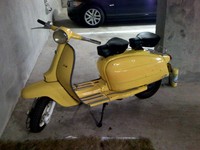}&\\[0pt]
&&(1, 1.00)&(2, 1.00)&(4, 1.00)&(5, 1.00)&(50, 1.00)&(100, 1.00)\\ & &
\includegraphics[width=38pt,height=38pt]{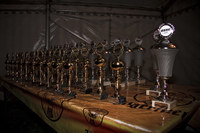}&
\includegraphics[width=38pt,height=38pt]{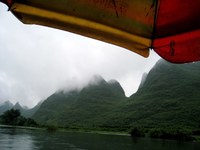}&
\includegraphics[width=38pt,height=38pt]{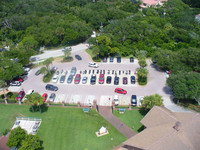}&
\includegraphics[width=38pt,height=38pt]{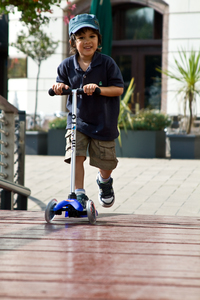}&
\includegraphics[width=38pt,height=38pt]{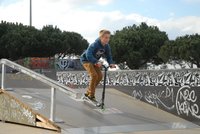}&
\includegraphics[width=38pt,height=38pt]{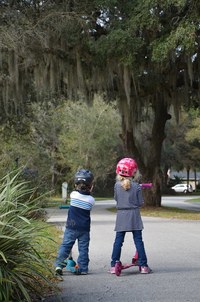}&\\[2pt]
\multicolumn{2}{c}{\scriptsize{motor scooter: \alert{84.8}$\rightarrow$\alert{94.0}}}&(18852,0.00)&(18853,0.00)&(18854,0.00)&(18855,0.00)&(18856,0.00)&(18857,0.00)\\[13pt]
%
%
\end{tabular}

%% file: experiments.tex
\section{Experiments}
\label{sec:exp}

\subsection{Experimental setup}
\label{sec:sec:evalproto}

\head{Datasets and task setup.}
We extend the \emph{Low-Shot ImageNet benchmark}~\cite{HG17}
by assuming many noisy examples in addition to the few clean ones.
In this benchmark, the 1000 ImageNet classes~\cite{russakovsky2015imagenet} are split into 389 base classes and 611 novel classes. The validation set contains 193 base and 300 novel classes, and the test set the remaining 196 base and 311 novel classes.
The base classes are used to learn the feature extractor (see supplementary material), while the novel classes form the set of classes $C$ on which we apply the cleaning and learn the classifier. We only assume noisy examples for the novel classes, not for the base ones.
Additionally, we apply a similar setup to the Places365 dataset~\cite{ZLK+17}.
We randomly choose 183 test and 182 validation classes.
We use the model learned on the base classes of Low-Shot ImageNet benchmark as the feature extractor.
Therefore, all classes in Places365 dataset are considered \textit{novel}.
We refer to this setup as \emph{Low-Shot Places365 benchmark}.

The standard benchmark includes $k$-shot classification, \ie classification on $k$ clean examples per class,
with $k \in \{1,2,5,10,20\}$. We extend it to $k$ clean and many noisy examples per class.
Similar to the work of Hariharan and Girshick~\cite{HG17}, we perform 5 episodes, each drawing a  subset of $k$ clean examples per class.
We report the average top-5 accuracy over the 5 episodes on novel \text{classes} of the test set.
Accuracy over all classes (base and novel) is reported in supplementary material.

\begin{figure}[t]
\input{fig_stats}
\caption{Noisy data statistics for Low-Shot ImageNet(top) and Low-Shot Places365(bottom). (a) Number of additional images collected from YFCC-100M per class $c$ for all novel classes. (b) Number of classes per group, when groups are created according to $|X_\cN^c|$ in logarithmic scale. (c) Accuracy improvement $\Delta\operatorname{Acc}$ (difference of accuracy between our method with noisy examples and the baseline without noisy examples) for prototype classifier, for same groups as in (b).\label{fig:stats}}
\end{figure}
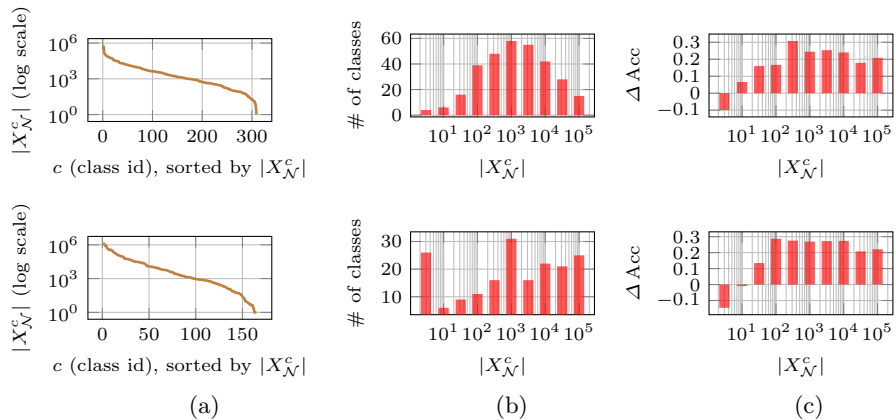

\head{Noisy data and statistics.}
We use the YFCC100M dataset~\cite{TSF+15} as a source of additional data with noisy labels. It contains approximatively 100M images collected from Flickr. Each image comes with a text description obtained from the user title and caption. We use the text description to obtain
images with noisy labels,
as discussed in Section~\ref{sec:problem}.

Figure~\ref{fig:stats} (top) shows the statistics of noisy examples for Low-Shot ImageNet benchmark.
The noisy examples for novel classes are long tailed in log scale (a).
Noisy examples per class differ significantly for different classes, with a minimum of zero for classes \emph{maillot} and \emph{missile}, and a maximum of 620,142 for the class \emph{church/church building}.
There is a significant number of classes where we obtain less than 1000 extra examples, but we improve
nevertheless; see Figure~\ref{fig:stats} (c). A small exception is 4 very rare classes out of 311, with around 3 additional images per class (leftmost bin in Figure~\ref{fig:stats} (b) and (c)). One could use more resources like web queries to collect additional data in real world applications.

Figure~\ref{fig:stats} (bottom) presents the same statistics for Low-Shot Places365 benchmark.
The trends are similar to those from Figure~\ref{fig:stats}, except that there are more than 20 classes without any additional noisy data in this task (b).
Nevertheless, there is a more consistent improvement in accuracy for classes that do have sufficient noisy data(c).

\begin{figure}[t]
\input{fig_cont_lambda}
\caption{Impact of $\lambda$ on the validation set of the extended Low-shot ImageNet benchmark with YFCC-100M for noisy examples using class prototypes~\eqref{equ:proto}.\label{fig:fig_cont_lambda}}
 \vspace{-10pt}
\end{figure}
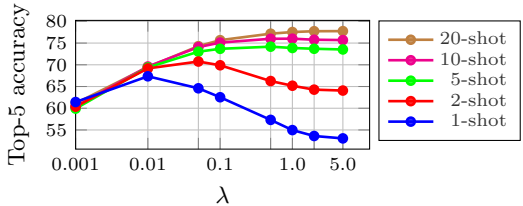

\head{Representation and classifier learning.}
In most experiments, we use ResNet-10~\cite{HZRS16} as feature extractor as in~\cite{GK18}.
Classification for novel classes is performed with \emph{class prototypes}~\eqref{equ:proto}, \emph{cosine classifier learning}~\eqref{equ:loss} or  \emph{deep network fine-tuning}. Hyper-parameters, such as batch size and number of epochs, are tuned on the validation set. We cross-validate the possible values of 512, 1024, 2048, 4096, and 8192 for batchsize and 10, 20, 30 and 50 for number of epochs. The learning rate starts from 0.1 and is reduced to 0.001 at the end of training with \emph{cosine annealing}~\cite{LH16}.
We handle the imbalance of the noisy set by normalizing by $r(X_c)$ in~\equ{loss}. Prototypes~\equ{proto} are used to initialize the weights $W$ of the cosine classifier in~\equ{loss}. We ignore examples $x_i$ with relevance $r(x_i) < 0.1$ to reduce the complexity when fine-tuning the network.
We also report results with ResNet-50 as feature extractor, using the model trained on base classes by~\cite{HG17}.
Following~\cite{DSH+18}, we apply PCA to the features to reduce their dimensionality to 256.
Base classes are represented by class prototypes~\equ{proto} in this case.

GCN training is performed with the Adam optimizer and a learning rate of $0.1$ for $100$ iterations.
We use dropout with probability 0.5. The dimensionality of the input descriptors is $d=512$ for ResNet-10 and $d=256$ for ResNet-50 (after PCA). Dimensionality of the internal representation in~\equ{gcnlayer} is $m=16$.
The affinity matrix is constructed with reciprocal top-50 nearest neighbors.

\head{Baseline methods.}
We implement and evaluate several baseline methods:
\begin{enumerate}[noitemsep,topsep=0pt]
\item \emph{$\beta$-cleaning} assigns $r(x_i) = \beta$ to all additional examples. We report results for $\beta=1.0$ (unit relevance score) and $\beta^{*}$, the optimal $\beta$ for all $k$ obtained on the validation set.
\item \emph{Similarity} uses the scaled cosine similarity as the relevance weight, \ie $r(x_i) = ( 1 +\vv_i^{\top} \vx) / 2$, where $\vx$ is the class prototype created with the features of clean examples.
\item \emph{Linear} learns a linear binary classifier where the positive instances are the $k$ labeled examples, and the negative examples are chosen randomly from other classes.
\item \emph{Label Propagation} (LP)~\cite{ZBL+03} propagates information by a linear operation. It solves the linear system $(I-\alpha D^{-1/2}AD^{-1/2}) \vr_c = \vy_c$~\cite{ITA+17} for each class $c$, where $D$ is the degree matrix of $A$, $\alpha = 0.9$  and $\vy_c \in \real^N$ is a $k$-hot binary vector indicating the clean (labeled) examples of class $c$. Relevance $r(x_i)$ is the $i$-th element $(\vr_c)_i$ of the solution.
\item \emph{MLP}, discussed in Section~\ref{sec:gcn}, learns a nonlinear mapping to assign relevance weights, but does not propagate over the graph. It is trained using~\eqref{equ:gcn-loss} and therefore includes part of our contribution.
\end{enumerate}

\begin{table}[t]
\input{abl_table_half}
\setlength{\tabcolsep}{1.2mm}
\caption{Comparison with baselines using noisy examples on the Low-shot ImageNet benchmark. We report top-5 accuracy on novel classes
with classification by class prototypes~\equ{proto}.
\label{tab:abl}
}
 \vspace{-10pt}
\end{table}

\subsection{Experimental results}
\head{The impact of the importance weight $\lambda$} is measured on the validation set and the best performing value is used on the test set for each value of $k$. Results on Low-shot ImageNet benchmark are shown in Figure~\ref{fig:fig_cont_lambda}. The larger the value of $\lambda$, the more the loss encourages noisy examples to be classified as negatives. As a consequence, large (small) $\lambda$ results in smaller (larger) relevance, on average, for noisy examples. The optimal $\lambda$ per value of $k$ suggests that 
as the number of clean examples decreases, the need for additional noisy ones increases.

\begin{table}[t]
\mbox{}\vspace{-.0cm}\\
\input{soa_table}
\setlength{\tabcolsep}{1.2mm}
\caption{Comparison to the state of the art on the Low-shot ImageNet benchmark. We report top-5 accuracy on novel classes.
We use class prototypes~\eqref{equ:proto}, cosine classifier learning~\eqref{equ:loss} and deep network fine-tuning for classification with our GCN-based data addition method.
$\dagger$ denotes numbers taken from the corresponding papers.
All other experiments are re-implemented by us.
\label{tab:soa}
 \vspace{-15pt}
}
\end{table}

\head{Comparison with baselines using additional data} on Low-shot Imagenet benchmark is presented in Table~\ref{tab:abl}.
Qualitative results are presented in Figure~\ref{fig:example2}.
The use of additional data is mostly harmful for $\beta$-weighting except for 1 and 2-shot.
MLP offers improvements in most cases, implying that it manages to appropriately down-weight irrelevant examples. The consistent improvement of GCN compared to MLP, especially large for small $k$, suggests that it is beneficial to incorporate relations, with the affinity matrix $A$ modeling the structure of the feature space. LP is a classic approach that also uses $A$ but is a linear operation with no parameters, and is inferior to our method.
The gain of cleaning ($\beta=1$ \vs ours) ranges from 11\% to 20\%.

\head{Comparison with the state of the art} on Low-shot Imagenet benchmark is presented in Table~\ref{tab:soa}. We significantly improve the performance by using additional data and cleaning compared to a number of different approaches, including the work by Gidaris and Komodakis~\cite{GK18}, which is our starting point. As expected, the gain is more pronounced for small $k$, reaching more than 20\% improvement for 1-shot novel accuracy.

Closest to ours is the work by Douze \etal~\cite{DSH+18}, who use the same experimental setup and the same additional data, but without filtering by text or using noisy labels. We outperform this approach in all cases, while requiring much less computation: \emph{offline}, we construct a separate small graph per class rather than a single graph over the entire 100M collection; \emph{online}, we perform inference by cosine similarity to one prototype per class or a learned classifier rather than iterative diffusion on the entire collection.
By ignoring examples that are not given any noisy label, we are only using a tiny fraction of the 100M collection: in particular, only 3,744,994 images for the 311-class test split of the Low-shot ImageNet benchmark.
In contrast to~\cite{DSH+18}, additional data brings improvement even at 20-shot with classifier learning or network fine-tuning.
Most importantly, our approach does not require the entire 100M collection at inference.

\head{Analysis of relevance weights.} 
We manually label all the noisy examples from 20 classes in order to quantitatively measure the accuracy of the assigned relevance.
We measure the noise ratio per class, \ie the ratio of irrelevant (negative) noisy images to all noisy (positive and negative) images. Positive and negative images are defined according to the manual labels. The 20 classes are selected such that 10 of them 
have
the highest $1$-shot accuracy, and the rest 
have
the lowest.
This allows us to examine 
success and failure cases.

In the case of $1$-shot classification ($k=1$), the average relevance weight is $0.71$ for positive and $0.40$ for negative examples. 
A success case is the ``bee eater'' class with noise ratio equal to $0.52$. 
Our method achieves $98.4\%$ accuracy for $1$-shot classification, compared to $68.8\%$ without any additional data. 
The average relevance weight is $0.99$ for positive examples and $0.25$ for negative examples of this class. 
One failure case is the ``muzzle'' class; it corresponds to the muzzle of an animal. 
The noise ratio is high; $94\%$ of the 980 collected images are not relevant with most being animals without a muzzle or a firearm. 
The 
1-shot classification 
accuracy
without noisy data is $4\%$. Our method offers only a small increase to $8\%$. 
This can be explained by inaccurate relevance weights, which are on average $0.18$ for positive and $0.30$ for negative examples.

\begin{table}[t]
\input{abl_table_places}
\setlength{\tabcolsep}{1.2mm}
\caption{Comparison with baselines using noisy examples on the Low-shot Places365 benchmark. We report top-5 accuracy on novel classes.
\label{tab:ablPlaces}
 \vspace{-15pt}
}
\end{table}

\head{Experiments in Low-Shot Places365} are reported in Table~\ref{tab:ablPlaces}.
Our results indicate that our method consistently outperforms the baselines on this benchmark as well.
Note that \textit{MLP}, which is also competitive for this task, is trained with our proposed loss function~\ref{equ:gcn-loss}. This is our contribution as well as the use of GCN.
These methods significantly improve over existing methods, such as Label Propagation~\cite{ZBL+03}.
Further improvements are brought by cosine classifier learning~\eqref{equ:loss} and deep network fine-tuning.

\begin{figure}[t]
\input{fig_example_places}

\vspace{-10pt}
\caption{Examples of clean images on Low-Shot Places365 Benchmark (left) for 1-shot classification, cumulative histogram of the predicted relevance for noisy images (middle), and representative noisy images (right), each having its position in the descending ranked list according to relevance value reported below. Test accuracy without and with additional data using prototypes~\equ{proto} is shown next to class names.
\label{fig:examplePlaces}
 \vspace{-10pt}
}
\end{figure}
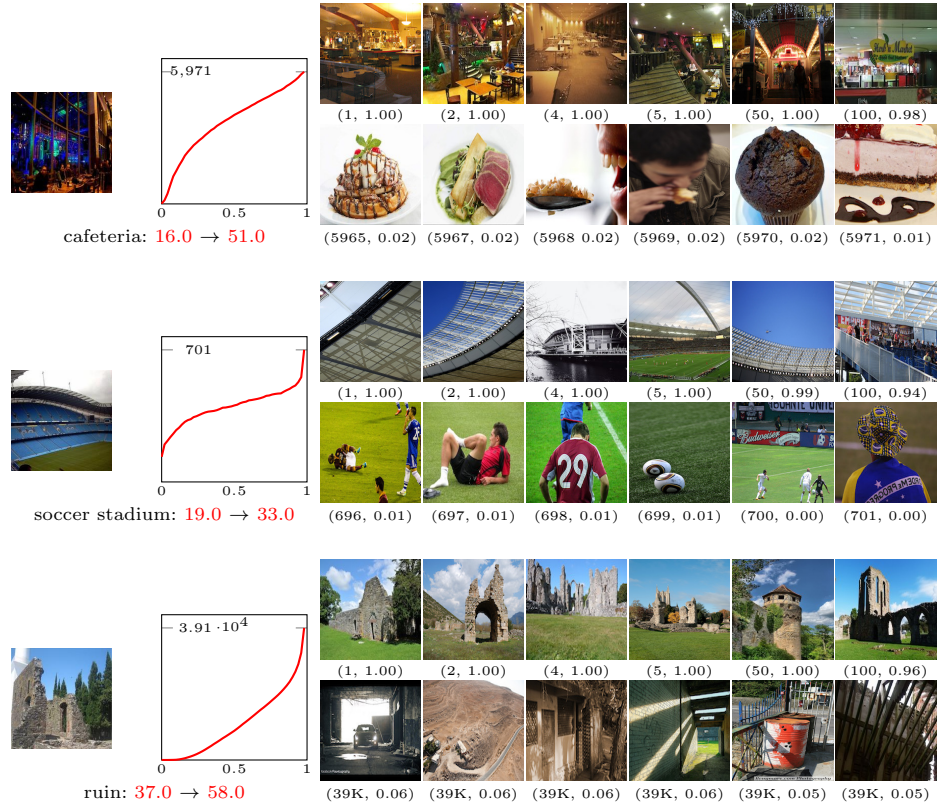

We also present qualitative results on Low-Shot Places365 in Figure~\ref{fig:examplePlaces}.
The first example at the top shows that top-ranked images by relevance depict different views of cafeterias for the \textit{cafeteria} class, while bottom-ranked images depict food served in a cafeteria, which are irrelevant to our task.
Similarly, our method assigns high relevance to images of soccer stadiums and low relevance to soccer players for the \textit{soccer} class.
Finally, our method finds similar images to the clean example for the \textit{ruin} class.
In general, top-ranking images exhibit diversity and are not just near-duplicates.

%% file: fig_stats.tex
\input{fig/data/sample}
\setlength{\tabcolsep}{3pt}
\begin{tabular}{ccc}
{
\begin{tikzpicture}
\begin{axis}[%
	width=0.33\linewidth,
	height=0.22\linewidth,
	xlabel = {$c$ (class id), sorted by $|X_\cN^c|$},
	ylabel = {$|X_\cN^c|$} (log scale),
	ymode = log
]
	\addplot[color=brown,     solid, no markers,  mark size=1.5, line width=1.0] table[x=sortimid, y expr={\thisrow{numim}}] \imperclass;
\end{axis}
\end{tikzpicture}
}
&
\begin{tikzpicture}
\begin{axis}[%
	ybar,
    xmode=log,
	width=0.33\linewidth,
	height=0.22\linewidth,
	xlabel = {$|X_\cN^c|$},
	ylabel = {\# of classes},
	minor tick style={draw=none},
    xtick={1,10,100,1000,10000,100000},
    xtick align=inside,
]
	\addplot[color=brown, fill = red,fill opacity=0.65,draw=none,bar width = 2] table[x=bincen, y expr={\thisrow{nclass}}] \difpernumim;
\end{axis}
\end{tikzpicture}
&
\begin{tikzpicture}
\begin{axis}[%
	ybar,
    xmode=log,
	width=0.33\linewidth,
	height=0.22\linewidth,
	xlabel = {$|X_\cN^c|$},
	ylabel = {$\Delta\operatorname{Acc}$},
	minor tick style={draw=none},
    xtick={1,10,100,1000,10000,100000},
    xtick align=inside,
]
	\addplot[color=brown, fill = red,fill opacity=0.65,draw=none,bar width = 2] table[x=bincen, y expr={\thisrow{mdif}}] \difpernumim;
\end{axis}
\end{tikzpicture}
\\
{
\begin{tikzpicture}
\begin{axis}[%
	width=0.33\linewidth,
	height=0.22\linewidth,
	xlabel = {$c$ (class id), sorted by $|X_\cN^c|$},
	ylabel = {$|X_\cN^c|$} (log scale),
	ymode = log
]
	\addplot[color=brown,     solid, no markers,  mark size=1.5, line width=1.0] table[x=sortimid, y expr={\thisrow{numim}}] \imperclassplaces;
\end{axis}
\end{tikzpicture}
}
&
\begin{tikzpicture}
\begin{axis}[%
	ybar,
    xmode=log,
	width=0.33\linewidth,
	height=0.22\linewidth,
	xlabel = {$|X_\cN^c|$},
	ylabel = {\# of classes},
	minor tick style={draw=none},
    xtick={1,10,100,1000,10000,100000},
    xtick align=inside,
]
	\addplot[color=brown, fill = red,fill opacity=0.65,draw=none,bar width = 2] table[x=bincen, y expr={\thisrow{nclass}}] \difpernumimplaces;
\end{axis}
\end{tikzpicture}
&
\begin{tikzpicture}
\begin{axis}[%
	ybar,
    xmode=log,
	width=0.33\linewidth,
	height=0.22\linewidth,
	xlabel = {$|X_\cN^c|$},
	ylabel = {$\Delta\operatorname{Acc}$},
	minor tick style={draw=none},
    xtick={1,10,100,1000,10000,100000},
    xtick align=inside,
]
	\addplot[color=brown, fill = red,fill opacity=0.65,draw=none,bar width = 2] table[x=bincen, y expr={\thisrow{mdif}}] \difpernumimplaces;
\end{axis}
\end{tikzpicture}
\\
~~~~~~~~~~~(a)&~~~~~~~~~~(b)&~~~~~~~~~~~~(c)\\
\end{tabular}

%% file: fig_cont_lambda.tex
\centering
\input{fig/data/sample}
{
\begin{tikzpicture}
\begin{axis}[%
	width=0.45\linewidth,
	height=0.27\linewidth,
	xlabel={\small $\lambda$},
   xmin = 0.001,
   	xtick={0.001,0.01,0.05,0.1,0.5,1.0,2.0,5.0},
   	xticklabels={0.001,0.01,,0.1,,1.0,,5.0},
   	xmode=log,
    grid=both,
	ylabel={\small Top-5 accuracy},
	legend cell align={left},
	legend pos=outer north east,
    legend style={cells={anchor=east}, font =\scriptsize, fill opacity=0.8, row sep=-2.5pt},
]

	\addplot[color=brown,     solid, mark=*,  mark size=1.5, line width=1.0] table[x=lambda, y expr={\thisrow{twenty}}] \yfccLambda;\leg{20-shot};
	\addplot[color=magenta,     solid, mark=*,  mark size=1.5, line width=1.0] table[x=lambda, y expr={\thisrow{ten}}] \yfccLambda;\leg{10-shot};
	\addplot[color=green,     solid, mark=*,  mark size=1.5, line width=1.0] table[x=lambda, y expr={\thisrow{five}}] \yfccLambda;\leg{5-shot};
	\addplot[color=red,     solid, mark=*,  mark size=1.5, line width=1.0] table[x=lambda, y expr={\thisrow{two}}] \yfccLambda;\leg{2-shot};
	\addplot[color=blue,     solid, mark=*,  mark size=1.5, line width=1.0] table[x=lambda, y expr={\thisrow{one}}] \yfccLambda;\leg{1-shot};
\end{axis}
\end{tikzpicture}
}

%% file: abl_table_half.tex
\centering
\scriptsize{
\begin{tabular}{@{\xssp}l@{\lsp}c@{\lsp}c@{\lsp}c@{\lsp}c@{\lsp}c}
\toprule
Method &$k$=1 & 2 & 5 & 10 & 20\\
\midrule
&\multicolumn{5}{l}{\textsc{Few Clean Examples}}\\
\midrule
Class proto.~\cite{GK18}& 45.3\std{0.65} & 57.1\std{0.37} & 69.3\std{0.32} & 74.8\std{0.20} & 77.8\std{0.24}\\
\midrule
&\multicolumn{5}{l}{\textsc{Few Clean \& Many Noisy Examples}}\\
\midrule
Similarity						& 49.8\std{0.29} & 56.3\std{0.27} & 64.2\std{0.32} & 68.4\std{0.14} & 71.2\std{0.12} \\
$\beta$-weighting, $\beta=1$	& 56.1\std{0.06} & 56.4\std{0.08} & 57.1\std{0.05} & 57.7\std{0.08} & 58.7\std{0.06} \\
$\beta$-weighting, $\beta^{*}$	& 55.6\std{0.24} & 58.3\std{0.14} & 63.4\std{0.25} & 67.5\std{0.34} & 71.0\std{0.22} \\
Linear							& 59.8\std{0.00} & 59.3\std{0.00} & 58.4\std{0.00} & 58.6\std{0.00} & 59.4\std{0.00} \\
Label Propagation				& 62.6\std{0.35} & 67.0\std{0.41} & 74.6\std{0.30} & 76.3\std{0.23} & 77.7\std{0.18} \\
MLP								& 63.6\std{0.41} & 68.8\std{0.42} & 73.7\std{0.25} & 75.6\std{0.21} & 77.6\std{0.21} \\
\midrule
Ours							& 67.8\std{0.10} & 70.9\std{0.30} & 73.9\std{0.17} & 76.1\std{0.12} & 78.2\std{0.14}\\
\hline
\end{tabular}
}

%% file: soa_table.tex
\centering
\scriptsize{
\begin{tabular}{@{\xssp}l@{\msp}l@{\msp}l@{\msp}l@{\msp}l@{\msp}l@{\xssp}}
\toprule
\textsc{Method} & \multicolumn{5}{c@{\hspace{8mm}}}{\textsc{Top-5 accuracy on novel classes}} \\
\midrule
&$k$=1 & 2 & 5 & 10 & 20\\
\midrule
\multicolumn{6}{c}{\textsc{ResNet-10~~--~~Few Clean Examples}}  \\ \midrule
Proto.-Nets~\cite{SSZ17}$^{\dagger}$ & 39.3 & 54.4 & 66.3 & 71.2 & 73.9 \\
Logistic reg. w/ H~\cite{WGH+18}$^{\dagger}$ & 40.7 & 50.8 & 62.0 & 69.3 & 76.5 \\
PMN w/ H~\cite{WGH+18}$^{\dagger}$ & 45.8 & 57.8 & 69.0 & 74.3 & 77.4 \\
Class proto.~\cite{GK18}& 45.3\std{0.65} & 57.1\std{0.37} & 69.3\std{0.32} & 74.8\std{0.20} & 77.8\std{0.24} \\
Class proto. w/ Att.~\cite{GK18}& 45.8\std{0.74} & 57.4\std{0.38} & 69.6\std{0.27} & 75.0\std{0.29} & 78.2\std{0.23} \\ \midrule
\multicolumn{6}{c}{\textsc{ResNet-10~~--~~Few Clean \& Many Noisy Examples} }  \\ \midrule
Ours - class proto.~\eqref{equ:proto}& \textbf{67.8\std{0.10}} & \textbf{70.9\std{0.30}} & \textbf{73.7\std{0.20}} & \textbf{76.1\std{0.16}} & \textbf{78.2\std{0.14}} \\
Ours - cosine~\eqref{equ:loss}& \textbf{73.2\std{0.14}} & \textbf{75.3\std{0.25}} & \textbf{75.6\std{0.24}} & \textbf{78.5\std{0.32}} & \textbf{80.7\std{0.26}} \\
Ours - fine-tune & \textbf{74.1\std{0.19}} & \textbf{76.2\std{0.28}} & \textbf{77.7\std{0.23}} & \textbf{80.6\std{0.31}} & \textbf{82.6\std{0.24}} \\
\midrule
\multicolumn{6}{c}{\textsc{ResNet-50~~--~~Few Clean Examples}}  \\ \midrule
Proto.-Nets~\cite{SSZ17}$^{\dagger}$ & 49.6 & 64.0 & 74.4 & 78.1 & 80.0 \\
PMN w/ H~\cite{WGH+18}$^{\dagger}$ & 54.7 & 66.8 & 77.4 & 81.4 & 83.8  \\ \midrule
\multicolumn{6}{c}{\textsc{ResNet-50~~--~~Few Clean \& Many Unlabeled Examples}}  \\ \midrule
Diffusion~\cite{DSH+18}$^{\dagger}$ 				& 63.6\std{0.61} & 69.5\std{0.60} & 75.2\std{0.40} & 78.5\std{0.34} & 80.8\std{0.18} \\
Diffusion - logistic~\cite{DSH+18}$^{\dagger}$ 	& 64.0\std{0.70} & 71.1\std{0.82} & 79.7\std{0.38} & 83.9\std{0.10} & 86.3\std{0.17} \\ \midrule
\multicolumn{6}{c}{\textsc{ResNet-50~~--~~Few Clean \& Many Noisy Examples}}  \\ \midrule
Ours - class proto.~\eqref{equ:proto}	& \textbf{69.7\std{0.44}} & \textbf{73.7\std{0.56}} & \textbf{77.0\std{0.20}} & \textbf{79.9\std{0.30}} & \textbf{81.9\std{0.29}} \\
Ours - cosine~\eqref{equ:loss}& \textbf{78.0\std{0.38}} & \textbf{80.2\std{0.33}} & \textbf{80.9\std{0.17}} & \textbf{83.7\std{0.19}} & \textbf{85.7\std{0.11}}  \\
Ours - fine-tune & \textbf{80.2\std{0.33}} & \textbf{82.6\std{0.14}} & \textbf{83.3\std{0.26}} & \textbf{85.9\std{0.22}} & \textbf{88.3\std{0.21}} \\
\bottomrule
\end{tabular}
}

%% file: abl_table_places.tex
\centering
\scriptsize{
\begin{tabular}{@{\xssp}l@{\lsp}c@{\lsp}c@{\lsp}c@{\lsp}c@{\lsp}c}
\toprule
Method &$k$=1 & 2 & 5 & 10 & 20\\
\midrule
&\multicolumn{5}{l}{\textsc{Few Clean Examples}}\\
\midrule
Class proto.~\cite{GK18}& 28.7\std{1.12} & 38.0\std{0.37} & 50.5\std{0.51} & 57.9\std{0.35} & 62.3\std{0.25}\\
\midrule
\multicolumn{6}{c}{\textsc{Few Clean \& Many Noisy Examples - Class proto.~\eqref{equ:proto}}}  \\ 
\midrule
$\beta$-weighting, $\beta=1$			& 44.0\std{0.34} & 45.7\std{0.22} & 48.4\std{0.31} & 50.0\std{0.12} & 50.8\std{0.25} \\
Label Propagation						& 39.6\std{0.78} & 46.5\std{0.22} & 54.8\std{0.42} & 59.6\std{0.11} & 62.0\std{0.14} \\
MLP										& 46.9\std{0.78} & 50.1\std{0.38} & 55.4\std{0.29} & 59.2\std{0.26} & 61.5\std{0.31} \\
Ours 									& 47.1\std{0.70} & 50.5\std{0.31} & 55.1\std{0.50} & 59.0\std{0.32} & 61.9\std{0.22}\\ 
\midrule
\multicolumn{6}{c}{\textsc{Few Clean \& Many Noisy Examples - Other classifiers}}  \\ 
\midrule
Ours - cosine~\eqref{equ:loss}			& 50.7\std{0.61} & 53.5\std{0.49} & 57.0\std{0.54} & 59.8\std{0.22} & 62.3\std{0.12}\\
Ours - fine-tune						& 51.8\std{0.69} & 54.8\std{0.57} & 59.5\std{0.63} & 62.9\std{0.39} & 66.0\std{0.27}\\
\bottomrule
\end{tabular}
}

%% file: fig_example_places.tex
\tiny
\begin{tabular}{@{\xssp}c@{\hspace{-50pt}}c@{\xssp}c@{\xssp}c@{\xssp}c@{\xssp}c@{\xssp}c@{\xssp}c@{\xssp}c@{\xssp}c@{\xssp}}
\multirow{2}{*}{\includegraphics[width=38pt,height=38pt]{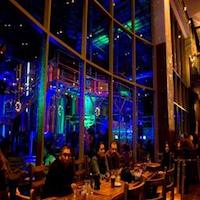}}&
\raisebox{13pt}{
\multirow{2}{*}{\input{fig/data/sample.tex}
\hspace{-13pt}
\begin{tikzpicture}
\begin{axis}[%
    width=100pt,
    height=100pt,
    xmin = 0, xmax = 1,
    xtick style={draw=none},
    ymin = 0,
    grid=none,
    ytick={5971},
    xtick = {0,0.5,1},    
    xticklabel style = {yshift=0.3ex,font=\tiny},
    yticklabel style = {xshift=10ex,font=\tiny},
    ylabel style = {yshift=-1.9ex,font=\tiny},
]
\addplot[color=red] table[x=x,y=z] \histclassthirtynineplaces;
\end{axis}
\end{tikzpicture}}}&
\includegraphics[width=38pt,height=38pt]{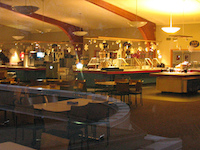}&
\includegraphics[width=38pt,height=38pt]{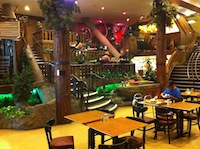}&
\includegraphics[width=38pt,height=38pt]{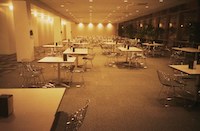}&
\includegraphics[width=38pt,height=38pt]{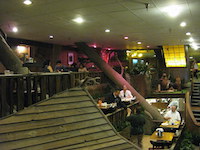}&
\includegraphics[width=38pt,height=38pt]{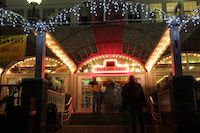}&
\includegraphics[width=38pt,height=38pt]{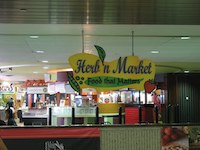}&\\[0pt]
&&(1, 1.00)&(2, 1.00)&(4, 1.00)&(5, 1.00)&(50, 1.00)&(100, 0.98)\\ & &
\includegraphics[width=38pt,height=38pt]{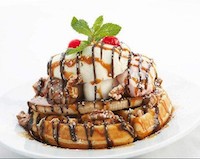}&
\includegraphics[width=38pt,height=38pt]{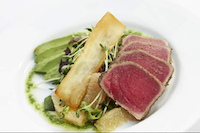}&
\includegraphics[width=38pt,height=38pt]{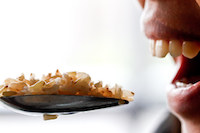}&
\includegraphics[width=38pt,height=38pt]{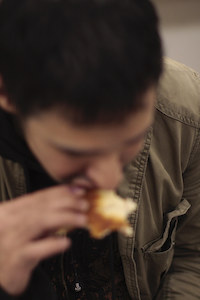}&
\includegraphics[width=38pt,height=38pt]{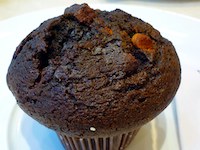}&
\includegraphics[width=38pt,height=38pt]{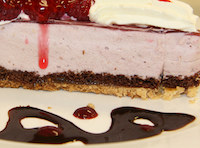}&\\[0pt]
\multicolumn{2}{c}{\scriptsize{cafeteria: \alert{16.0} $\rightarrow$ \alert{51.0} }}&(5965, 0.02)&(5967, 0.02)&(5968 0.02)&(5969, 0.02)&(5970, 0.02)&(5971, 0.01)\\[13pt]
%
%
\multirow{2}{*}{\includegraphics[width=38pt,height=38pt]{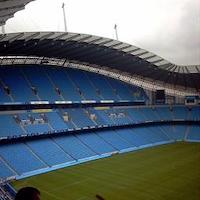}}&
\raisebox{13pt}{
\multirow{2}{*}{\input{fig/data/sample.tex}
\hspace{-13pt}
\begin{tikzpicture}
\begin{axis}[%
    width=100pt,
    height=100pt,
    xmin = 0, xmax = 1,
    xtick style={draw=none},
    ymin = 0,
    grid=none,
    ytick={701},
    xtick = {0,0.5,1},    
    xticklabel style = {yshift=0.3ex,font=\tiny},
    yticklabel style = {xshift=10ex,font=\tiny},
    ylabel style = {yshift=-1.9ex,font=\tiny},
]
\addplot[color=red] table[x=x,y=z] \histclassonehundredfiftyplaces;
\end{axis}
\end{tikzpicture}}}&
\includegraphics[width=38pt,height=38pt]{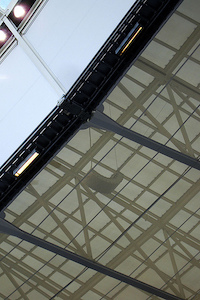}&
\includegraphics[width=38pt,height=38pt]{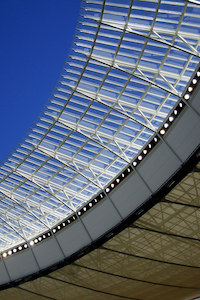}&
\includegraphics[width=38pt,height=38pt]{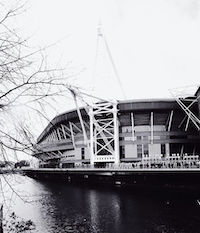}&
\includegraphics[width=38pt,height=38pt]{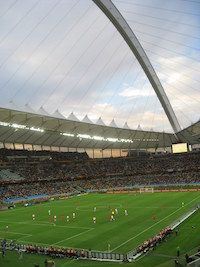}&
\includegraphics[width=38pt,height=38pt]{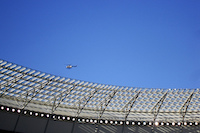}&
\includegraphics[width=38pt,height=38pt]{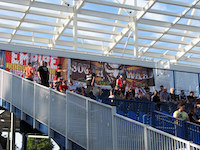}&\\[0pt]
&&(1, 1.00)&(2, 1.00)&(4, 1.00)&(5, 1.00)&(50, 0.99)&(100, 0.94)\\ & &
\includegraphics[width=38pt,height=38pt]{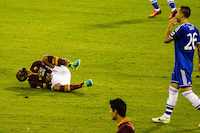}&
\includegraphics[width=38pt,height=38pt]{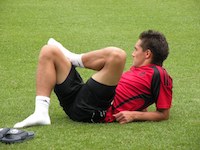}&
\includegraphics[width=38pt,height=38pt]{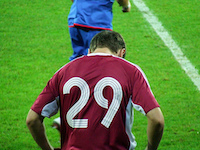}&
\includegraphics[width=38pt,height=38pt]{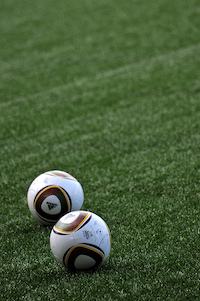}&
\includegraphics[width=38pt,height=38pt]{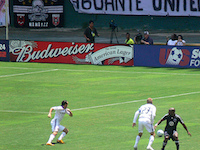}&
\includegraphics[width=38pt,height=38pt]{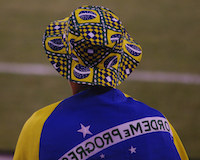}&\\[0pt]
\multicolumn{2}{c}{\scriptsize{soccer stadium: \alert{19.0} $\rightarrow$ \alert{33.0} }}&(696, 0.01)&(697, 0.01)&(698, 0.01)&(699, 0.01)&(700, 0.00)&(701, 0.00)\\[13pt]
%
%
\multirow{2}{*}{\includegraphics[width=38pt,height=38pt]{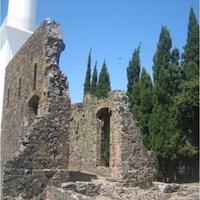}}&
\raisebox{13pt}{
\multirow{2}{*}{\input{fig/data/sample.tex}
\hspace{-13pt}
\begin{tikzpicture}
\begin{axis}[%
    width=100pt,
    height=100pt,
    xmin = 0, xmax = 1,
    xtick style={draw=none},
    ymin = 0,
    grid=none,
    ytick={39120},
    xtick = {0,0.5,1},    
    xticklabel style = {yshift=0.3ex,font=\tiny},
    yticklabel style = {xshift=10ex,font=\tiny},
    ylabel style = {yshift=-1.9ex,font=\tiny},
    every y tick scale label/.style={at={(yticklabel cs:0.935,5pt)},xshift=2.5em,left,inner sep=0pt}
]
\addplot[color=red] table[x=x,y=z] \histclassonehundredfortythreeplaces;
\end{axis}
\end{tikzpicture}}}&
\includegraphics[width=38pt,height=38pt]{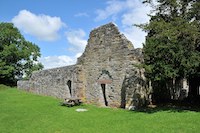}&
\includegraphics[width=38pt,height=38pt]{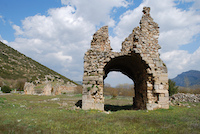}&
\includegraphics[width=38pt,height=38pt]{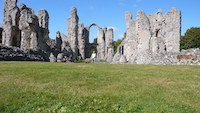}&
\includegraphics[width=38pt,height=38pt]{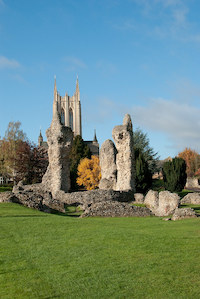}&
\includegraphics[width=38pt,height=38pt]{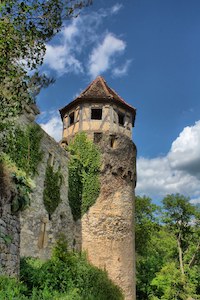}&
\includegraphics[width=38pt,height=38pt]{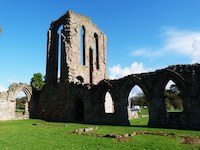}&\\[0pt]
&&(1, 1.00)&(2, 1.00)&(4, 1.00)&(5, 1.00)&(50, 1.00)&(100, 0.96)\\ & &
\includegraphics[width=38pt,height=38pt]{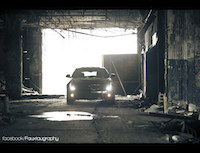}&
\includegraphics[width=38pt,height=38pt]{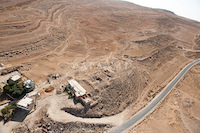}&
\includegraphics[width=38pt,height=38pt]{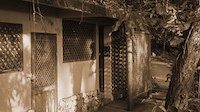}&
\includegraphics[width=38pt,height=38pt]{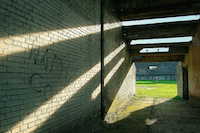}&
\includegraphics[width=38pt,height=38pt]{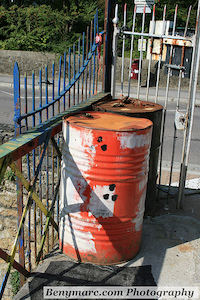}&
\includegraphics[width=38pt,height=38pt]{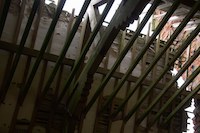}&\\[0pt]
\multicolumn{2}{c}{\scriptsize{ruin: \alert{37.0} $\rightarrow$ \alert{58.0}} }&(39K, 0.06)&(39K, 0.06)&(39K, 0.06)&(39K, 0.06)&(39K, 0.05)&(39K, 0.05)\\[13pt]
%
%
\end{tabular}

%% file: conclusions.tex
\section{Conclusions}
\label{sec:conclusions}

In this paper we have introduced a new method for assigning class relevance to noisy images obtained by textual queries with class names.
Our approach leverages one or a few labeled images per class and
relies on a graph convolutional network (GCN) to propagate visual information from the labeled images to the noisy ones. The GCN is trained as a binary classifier discriminating clean from noisy examples using a weighted binary cross-entropy loss function and inferring ``clean'' probability as a relevance score for that class. Experimental results show that using noisy images weighted by this relevance score significantly improves the classification accuracy.

%% file: appendix.tex
\appendix
\section{The role of base classes}
\label{sec:appendix_a}
The proposed method is applicable with any given feature extractor. Herein, we describe the learning of the feature extractor on a set of base classes according to a standard few-shot learning setup and benchmark~\cite{HG17}. Then, we describe the extended classifiers to the union of all classes, \ie base classes and novel classes, which are the ones used in Section~\ref{sec:evalproto}.

\subsection{Representation learning on base classes}
We are given a set $X_\cB \subset \cX$ of examples, each having a clean label in a set of \emph{base classes} $C_\cB$ with $\card{C_\cB} = K_\cB$.
Base classes $C_\cB$ are disjoint from $C$, which are also known as novel classes.
These data are used to learn a feature representation, \ie a feature extractor $g_\theta$, by learning a $K_\cB$-way base-class classifier for unseen data in $\cX$.
The parameters $\theta$ of the feature extractor and $W_\cB$ of the classifier are jointly learned by minimizing the cross entropy loss
\begin{equation}
L_\cB(C_\cB, X_\cB; \theta, W_\cB) = -\sum_{c \in C_\cB} \frac{1}{\card{X_\cB^c}} \sum_{x \in X_\cB^c}
	\log(\vsigma(s \hat{W}_\cB\tran \hat{g}_\theta(x))_c).
\label{equ:base-loss}
\end{equation}
The learned feature extractor parameters $\theta$ and the learned scale parameter $s$ are used by our method as described Sections~\ref{sec:gcn} and~\ref{sec:evalproto}.

\subsection{Classification on all classes}
The classifier parameters $W_\cB$ are used, combined with classifier parameters $W$ learned as described in Section~\ref{sec:evalproto}, for classification on \emph{all classes} $C_\cA = C \cup C_\cB$.

\head{Class prototypes.}
The concatenated parameter matrix $W_\cA = [W_\cB, W]$ is used for $K_\cA$-way prediction on all (base and novel) classes by $\pi_{\theta,W_\cA}$, where $K_\cA = K + K_\cB$. $W_\cB$ is learned according to $L_\cB(C_\cB, X_\cB; \theta, W_\cB)$~\equ{base-loss}, while $W$ is learned according to
\equ{proto}.

\head{Cosine classifier learning.}
Prediction on all classes is made as in the previous case, but $W$ is learned according to \equ{loss}.

\head{Deep network fine-tuning.} We now assume that base class examples are accessible too and, given all examples $X_\cA = X_\cB \cup X_\cE$, we jointly learn the parameters $\theta$ of the feature extractor and $W_\cA = [W_\cB, W]$ of the $K_\cA$-way cosine classifier for all classes by minimizing loss function
\begin{equation}
L_\cA(C_\cA, X_\cA; \theta, W_\cA) = L_\cB(C_\cB, X_\cB; \theta, W_\cB) + L(C, X_\cE; \theta, W).
\label{equ:all-loss}
\end{equation}
Note that in contrast to \equ{loss}, the last term of \equ{all-loss} optimizes parameters $\theta$ too.
 As mentioned earlier, such learning is typically avoided in a few-shot learning setup.
In few cases, it takes the form of fine-tuning including all base class data~\cite{Qi_2018_CVPR}, or only lasts for a few iterations when the base class data is not accessible~\cite{finn2017model}.

\subsection{Results on all classes}

\begin{table*}
\mbox{}\vspace{-.0cm}\\
\input{soa_table_all}

\setlength{\tabcolsep}{1.2mm}
\caption{Comparison to the state of the art on the Low-shot ImageNet benchmark. We report top-5 accuracy on all classes.
We use class prototypes~\eqref{equ:proto}, cosine classifier learning~\eqref{equ:loss} and deep network fine-tuning for classification with our GCN-based data addition method.
$\dagger$ denotes numbers taken from the corresponding papers.
All other experiments are re-implemented by us.
\label{tab:soaAll}
}
\end{table*}

We report the accuracy over all classes in Table~\ref{tab:soaAll}. When fine-tuning the network by~\equ{all-loss},  the learned $W$ is used to initialize the corresponding part of $W_\cA$ and we train all layers for 10 epochs with learning rate 0.01.
The results indicate that our method still brings significant improvements when all classes are used.

\section{Results on Mini-Imagenet}

We evaluate the proposed method on another popular benchmark, \ie few-shot learning on Mini-ImageNet~\cite{VBL+16}.
The dataset is a subset of ImageNet~\cite{russakovsky2015imagenet}, and contains $100$ different classes, split into $64$ base, $16$ validation and $20$ test classes~\cite{RL16}.
Each class contains $600$ images that are re-sized to a resolution of $84 \times 84$.
We use the ConvNet-128 model with cosine classifier, following~\cite{GK18}.
Novel categories are classified using class prototypes~\eqref{equ:proto}.

\begin{table}
\input{abl_table_mini_imagenet}
\setlength{\tabcolsep}{1.2mm}
\caption{Comparison with baselines using noisy examples on the Mini-ImageNet dataset. We report the accuracy for $5$-way $k$-shot experiments where $k=1$ and $k=5$.
\label{tab:ablMini}
 \vspace{-10pt}
}
\end{table}

Table~\ref{tab:ablMini} shows the accuracy on Mini-Imagenet for the $5$-way $k$-shot classification scenario with $k=1$ and $k=5$.
We report the average accuracy over $600$ trials along with the confidence interval.
Our method brings significant improvements for $k=1$, showing its generalization across different few-shot datasets and benchmarks.

%% file: soa_table_all.tex
\centering
\tiny{
\resizebox{0.99\linewidth}{!}{
\begin{tabular}{@{\xssp}l@{\msp}l@{\msp}l@{\msp}l@{\msp}l@{\msp}l@{\xssp}}
\toprule
\textsc{Method} & \multicolumn{5}{c@{\hspace{8mm}}}{\textsc{Top-5 accuracy on all classes}} \\
\midrule
&$k$=1 & 2 & 5 & 10 & 20\\
\midrule
& \multicolumn{5}{l}{\textsc{ResNet-10~~--~~Few Clean Examples}}  \\ \midrule
Proto.-Nets~\cite{SSZ17}$^{\dagger}$ 			& 49.5 & 61.0 & 69.7 & 72.9 & 74.6 \\
Logistic reg. w/ H~\cite{WGH+18}$^{\dagger}$ 	& 54.4 & 61.0 & 69.0 & 73.7 & 76.5 \\
PMN w/ H~\cite{WGH+18}$^{\dagger}$ 				& 40.8 & 49.9 & 64.2 & 71.9 & 76.9 \\
Class proto.~\cite{GK18}			& 57.0\std{0.36} & 64.7\std{0.16} & 72.5\std{0.18} & 75.8\std{0.16} & 77.4\std{0.19} \\
Class proto. w/ Att.~\cite{GK18}	& 58.1\std{0.48} & 65.2\std{0.15} & 72.9\std{0.25} & 76.6\std{0.18} & 78.8\std{0.16} \\ \midrule
&\multicolumn{5}{l}{\textsc{ResNet-10~~--~~Few Clean \& Many Noisy Examples} }  \\ \midrule
Ours - class proto.~\eqref{equ:proto}& \textbf{70.3\std{0.05}} & \textbf{72.1\std{0.18}} & \textbf{74.1\std{0.12}} & \textbf{75.6\std{0.13}} & \textbf{76.9\std{0.09}}\\
Ours - cosine~\eqref{equ:loss}		& \textbf{72.4\std{0.07}} & \textbf{73.4\std{0.21}} & \textbf{77.2\std{0.20}} & \textbf{78.8\std{0.21}} & \textbf{79.2\std{0.17}}\\
Ours - fine-tune 					& \textbf{76.0\std{0.10}} & \textbf{77.3\std{0.13}} & \textbf{78.7\std{0.19}} & \textbf{80.7\std{0.25}} & \textbf{82.2\std{0.14}}\\
\midrule
&\multicolumn{5}{l}{\textsc{ResNet-50~~--~~Few Clean Examples}}  \\ \midrule
Proto.-Nets~\cite{SSZ17}$^{\dagger}$ 			& 61.4 & 71.4 & 78.0 & 80.0 & 81.1 \\
PMN w/ H~\cite{WGH+18} $^{\dagger}$				& 65.7 & 73.5 & 80.2 & 82.8 & 84.5  \\ \midrule
&\multicolumn{5}{l}{\textsc{ResNet-50~~--~~Few Clean \& Many Noisy Examples}}  \\ \midrule
Ours - class proto.~\eqref{equ:proto}		& \textbf{73.8\std{0.33}} & \textbf{76.6\std{0.36}} & \textbf{78.9\std{0.19}} & \textbf{80.8\std{0.21}} & \textbf{82.2\std{0.14}}\\
Ours - cosine~\eqref{equ:loss}				& \textbf{78.2\std{0.25}} & \textbf{79.6\std{0.23}} & \textbf{80.4\std{0.18}} & \textbf{82.4\std{0.19}} & \textbf{84.1\std{0.09}}  \\
Ours - fine-tune 							& \textbf{81.6\std{0.20}} & \textbf{83.2\std{0.16}} & \textbf{84.3\std{0.23}} & \textbf{86.2\std{0.17}} & \textbf{87.8\std{0.03}} \\
\bottomrule
\end{tabular}
}
}

%% file: abl_table_mini_imagenet.tex
\centering
\footnotesize{
\begin{tabular}{@{\xssp}l@{\msp}cc}
\toprule
Method &$k$=1 & $k$=5 \\
\midrule
&\multicolumn{2}{l}{\textsc{Few Clean Examples}}\\
\midrule
Class proto.~\cite{GK18}				& 54.2\std{0.77} & 71.2\std{0.61}	\\
Class proto. w/ Att.~\cite{GK18}		& 56.2\std{0.81} & 72.9\std{0.62}	\\
\midrule
\multicolumn{3}{c}{\textsc{Few Clean \& Many Noisy Examples - Class proto.~\eqref{equ:proto}}}  \\ 
\midrule
$\beta$-weighting, $\beta=1$			& 63.5\std{0.77} & 65.2\std{0.81} \\
Label Propagation						& 67.0\std{0.74} & 74.8\std{0.61} \\
MLP										& 65.9\std{0.78} & 73.9\std{0.63} \\
Ours 									& 68.2\std{0.76} & 74.7\std{0.59} \\ 
\bottomrule
\end{tabular}
}